\documentclass{article}

\usepackage[main,final]{neurips_2026}

\usepackage{microtype}
\usepackage{graphicx}
\usepackage{float}  %
\usepackage{subcaption}
\usepackage{booktabs} %
\usepackage{multirow}

\usepackage{hyperref}

\usepackage{amsmath}
\usepackage{amssymb}
\usepackage{mathtools}
\usepackage{amsthm}
\usepackage{amsfonts}

\usepackage[capitalize,noabbrev]{cleveref}

\theoremstyle{plain}

\theoremstyle{definition}

\theoremstyle{remark}

\usepackage{bm}
\usepackage{makecell}
\usepackage{tabularx}
\usepackage{algorithmic}
\usepackage{xcolor}
\usepackage{xspace}
\usepackage{listings}
\newcolumntype{Y}{>{\centering\arraybackslash}X}
\usepackage{arydshln} %
\usepackage{pifont}
\usepackage[most]{tcolorbox} %
\usepackage[utf8]{inputenc} %
\usepackage[T1]{fontenc}    %
\usepackage{url}            %
\usepackage{nicefrac}       %
\usepackage{wrapfig}

\newcommand{\method}{\textit{Diff-Aid}\xspace}
\newcommand{\module}{\textit{Aid}\xspace}

\title{\method: Inference-time Adaptive Interaction Denoising for Rectified Text-to-Image Generation}

\author{%
  Binglei Li\textsuperscript{1,2} \quad
  Mengping Yang\textsuperscript{3}\thanks{Project Lead.} \quad
  Zhiyu Tan\textsuperscript{1,3} \quad
  Qiang Xiang\textsuperscript{1} \\
  \bfseries Junping Zhang\textsuperscript{1}\thanks{Corresponding Authors.} \quad
  Hao Li\textsuperscript{1,2,3}\footnotemark[2] \\[0.6em]
  \textsuperscript{1}\,Fudan University \quad
  \textsuperscript{2}\,Shanghai Innovation Institute \quad
  \textsuperscript{3}\,Shanghai Academy of AI for Science \\
  \texttt{blli24@m.fudan.edu.cn},\quad \texttt{\{mengpingyang, jpzhang, lihao\_lh\}@fudan.edu.cn}
}

\makeatletter
\def\@fnsymbol#1{%
  \ensuremath{%
    \ifcase#1%
      \or \dagger%
      \or *      %
      \or \ddagger
      \or \mathsection
      \or \mathparagraph
      \or \|
      \or **%
      \or \dagger\dagger
      \or \ddagger\ddagger
    \fi
  }%
}
\makeatother

\begin{document}

\maketitle

\begin{figure}[H]
\vspace{-20pt}
\centering
\includegraphics[width=0.95\textwidth]{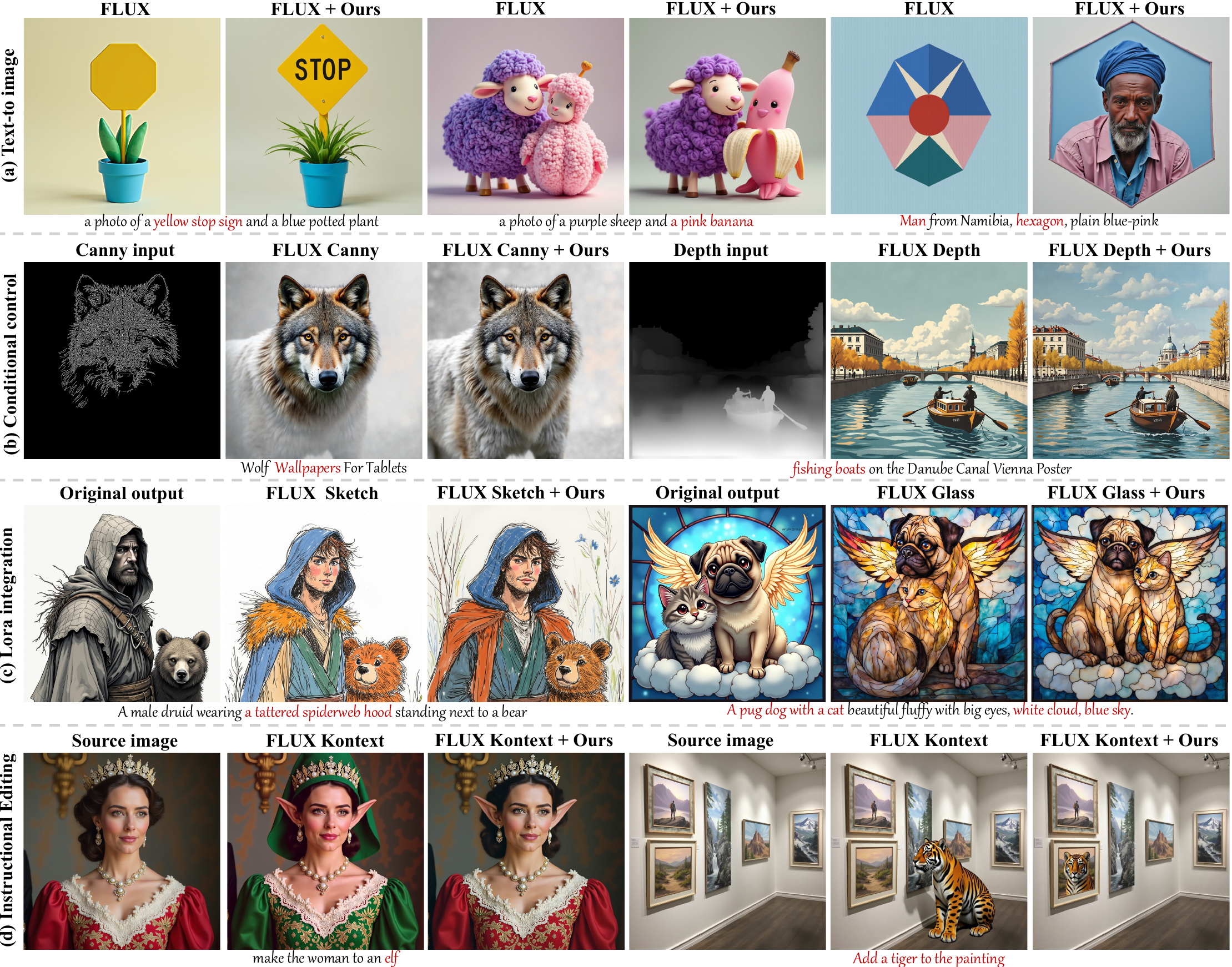}
\caption{
\textbf{
We present \method to adaptively improve the interaction between textual conditions and image latents during test time.
}
Benefited from our \textit{plug-in} and \textit{interpretable} design, \method enables improved prompts following and synthesis quality for (a) text-to-image generation,
(b) adding conditional input for controllable generation,
(c) integrating off-the-shelf LoRAs and (d) zero-shot instructional image editing.
}
\label{fig:teaser}
\vspace{-15pt}
\end{figure}

\begin{abstract}
Recent text-to-image (T2I) diffusion models have achieved remarkable advancement, yet faithfully following complex textual descriptions remains challenging due to insufficient interactions between textual and visual features.
Prior approaches enhance such interactions via architectural design or handcrafted textual condition weighting, but lack flexibility and overlook the dynamic interactions across different blocks and denoising stages.
To provide a more flexible and efficient solution, we propose \method, a lightweight module trained with a frozen backbone that predicts signed, per-token text modulation across transformer blocks and denoising timesteps at inference.
Beyond improving generation quality, \method yields interpretable modulation patterns that reveal how different blocks, timesteps, and textual tokens contribute to semantic alignment during denoising.
As a plug-and-play module, \method can be seamlessly integrated into downstream applications for further improvement, including style LoRAs, controllable generation, and zero-shot editing.
Experiments on SD~3.5 and FLUX show improvements in compositional prompt following, visual quality, and blinded human preference.
Our code is available at \href{https://github.com/phil329/Diff-Aid}{https://github.com/phil329/Diff-Aid}.

\end{abstract}

\vspace{-10pt}
\section{Introduction}
\vspace{-5pt}

\label{sec:intro}
Diffusion models (DMs)~\cite{ddpm_ho2020denoising,ddim_song2020}, which iteratively denoises random noisy samples into high-quality outputs, have revolutionized the experiences of AI generated content (AIGC) in various applications including text-to-image~\cite{ma2024latte,SD3_esser2024scaling}, text-to-video generation~\cite{yang2024cogvideox,hunyuan_video_2024,OpenAI2024_Sora,wan2025wan}, joint audio-video synthesis~\cite{wang2026klear, zhang2025uniavgen}, \emph{etc.}
These breakthroughs mainly stem from the scalability of diffusion transformers (DiT)~\cite{dit_AdaLN_peebles2023scalable,RectifiedFlow_liuflow} over conventional U-Net models~\cite{rombach2022high,diffusion_beat_gan_dhariwal2021}, as well as advances in large-scale training compute, infra and data pipelines.
Yet, considering DMs as a model that maps textual conditions to visual outputs, they might often fail to produce images that accurately reflect textual details due to insufficient interactions between textual conditions and image features.
For instance, the current top-performing model FLUX~\cite{flux_2024} failed to produce the ``\textit{yellow stop sign}" in Fig.~\ref{fig:teaser} (a).

Many prior approaches have been developed to strengthen cross-modal interactions between text and image representations.
On the architecture design hand, early methods rely on cross-attention mechanisms to inject textual information into diffusion models~\cite{rombach2022high,kolors_2024}, while DiT~ \cite{dit_AdaLN_peebles2023scalable} proposed adaLN-zero strategy to condition transformer blocks on class conditions.
The most recent multimodal diffusion transformers (MMDiT) based approaches~\cite{flux_2024,SD35_2024,qwenimage_wu2025technicalreport} employ a shared unified transformer to jointly model both modalities, facilitating cross-modal interactions.
However, training such models from scratch requires massive compute and data.
On the other hand, classifier-free guidance (CFG)~\cite{cfg_ho2022classifier} adopts a static scaling factor to balance conditional and unconditional model outputs, thus improving conditioning strength.
TACA~\cite{taca_lv2025rethinking} pointed out that semantic discrepancies may arise from the suppression of cross-modal attention due to the numerical asymmetry between the number of visual and text tokens and proposed a handcrafted weighting strategy to enhance textual tokens.
Further, Li et al.~\cite{li2026unraveling} probed specific roles of transformer blocks in rendering textual conditions and attached block-specific enhancements to improve interactions.
Yet, they require heuristic search of weighting strategies across the forward process, without explicitly considering the dynamic interactions between textual and visual features throughout the denoising process.

\begin{wrapfigure}{r}{0.5\textwidth}
    \centering
    \setlength{\abovecaptionskip}{3pt}
    \setlength{\belowcaptionskip}{-3pt}
    \includegraphics[width=\linewidth]{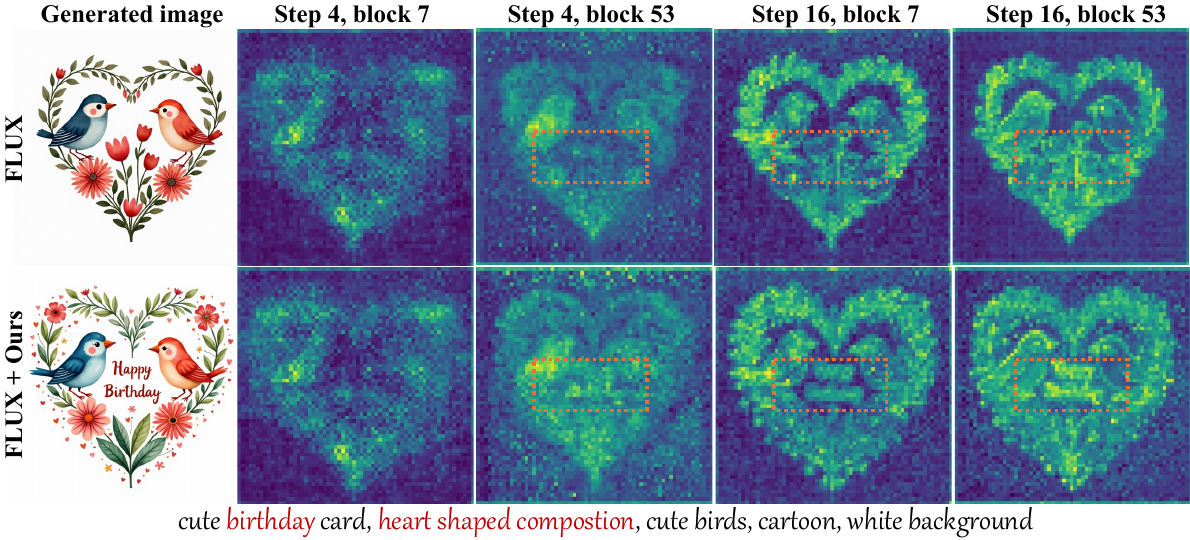}
    \caption{
    \textbf{Attention visualization of different blocks across denoising steps.}
    If the interaction between textual conditions and image latents is insufficient, the rendered image may fail to faithfully reflect given conditions.
    }
    \label{fig:attn_map_example}
    \vspace{-1em}
\end{wrapfigure}

In this paper, we aim to develop a lightweight, plug-in, and adaptive method to dynamically adjust interactions between textual and visual features during inference time, thereby delivering high-quality output images with improved prompt following and human preferences.
Note that \method requires a lightweight training phase to learn adaptive weights, which is far more efficient than retraining the entire model.
Fig.~\ref{fig:attn_map_example} illustrates a localized failure: vanilla FLUX omits ``birthday'' and shows a weak, spatially diffuse response to that token, whereas \method recovers the text and a more focused response from the same prompt and initial noise.
The block- and timestep-dependent attention patterns motivate an intervention before joint attention.
Because prompt tokens also differ in visual relevance, we learn token-specific modulation rather than applying one global conditioning strength.

Capitalizing on these observations, we propose \method, an inference-time \textbf{A}daptive \textbf{I}nteraction \textbf{D}enoising module for text-to-image \textbf{Diff}usion models. ``Inference-time'' describes its dynamic use during sampling: the module is first trained with the backbone frozen, then predicts modulation coefficients for each prompt and denoising step.
Concretely, \method introduces two key innovations: \textit{1)} an \module module, which explicitly adjusts textual conditions according to the varied blocks, timesteps, and tokens;
and \textit{2)} a gated sparsity and stabilized regularization mechanism that respectively ensure attaching more attention to important blocks and avoiding training collapse.
In this way, the learned parameters are interpretable with respect to different blocks, varied timesteps, and textual tokens, further providing insights into the model's internal behavior (as illustrated in Fig.~\ref{fig:alpha}).
More importantly, our design is lightweight with a small number of trainable parameters, without modifying the representation space of pretrained models.
As illustrated in Fig.~\ref{fig:teaser}, as a plug-in module, \method can be seamlessly integrated into homologous models for controllable image generation (b), styled LoRA generation (c), and instructional image editing (d), demonstrating its versatility and effectiveness.
We evaluate our \method on two strong baseline models (\emph{i.e.,} SD 3.5~\cite{SD3_esser2024scaling} and FLUX~\cite{flux_2024}), the consistent performance improvements across varied metrics and benchmarks demonstrate the effectiveness of our design.
Further analysis reveals that \method learns meaningful and interpretable interaction patterns aligned with the denoising dynamics, validating our design principles.

To sum up, our contributions are:
    \textit{1)} We propose \method, a lightweight model to adaptively manipulate the interactions between textual and visual features at inference-time, capturing the dynamic relationships across different blocks, timesteps, textual tokens, as well as the intrinsic properties of baseline models.
    \textit{2)} Our \method is designed as a plug-in and generalizable module that can be seamlessly integrated for downstream applications, including LoRAs, controllable generation, and zero-shot image editing.
    \textit{3)} Extensive experiments across multiple models and evaluation metrics demonstrate consistent improvements in prompt following, visual quality, and human preferences.

\vspace{-10pt}

\section{Methodology}
\label{sec:methods}

\vspace{-5pt}
\subsection{Motivation and Design Principles}
\label{sec:motivations}

\noindent \textbf{Interactions between textual conditions and image latents for T2I.}
Text is the primary guidance for image generation in text-to-image diffusion models.
The diffusion model is indeed a mapper $\mathcal{F}: \mathcal{P}_{txt} \rightarrow \mathcal{X}_{img}$, where the prompt is the only condition to generate the image.
MMDiT models~\cite{SD3_esser2024scaling,flux_2024} leverage joint self-attention to integrate text and image features:
\setlength{\abovedisplayskip}{3pt}
\setlength{\belowdisplayskip}{3pt}
\setlength{\abovedisplayshortskip}{2pt}
\setlength{\belowdisplayshortskip}{2pt}
\begin{equation}
    \label{eq:self_attention}
    \begin{aligned}
    &\mathbf{z}_{t}^{l+1} = \text{Softmax}\left(\mathbf{Q} \mathbf{K}^T/\sqrt{d}\right) \mathbf{V} + \mathbf{z}_{t}^{l}; \\
    &[\mathbf{Q}, \mathbf{K}, \mathbf{V}] = W_{qkv} \cdot \text{LN}\left([\mathbf{z}_{t}^{l}, \mathbf{c}_t^l]\right).
    \end{aligned}
\end{equation}
Therefore, effective interaction in attention between textual and visual tokens is vitally important for high-quality image generation.
Fig.~\ref{fig:attn_map_example} shows two denoising trajectories of the same prompt and initial noise but different text-image interaction patterns.
The attention maps of the second trajectory have a stronger focus on the relevant text token ``birthday'', leading to better text-image alignment.
\begin{tcolorbox}[colback=gray!10, colframe=gray!70, coltext=black, title=Principle 1: Cross-modal interaction matters during denoising,width=\linewidth, fontupper=\small]
    \vspace{-5pt}
    Attention mechanism enables text-image interaction, which is crucial for high-quality image generation in Diffusion models.
    Transformer blocks exhibit varying interaction patterns over denoising steps.
    \vspace{-5pt}
\end{tcolorbox}

\noindent \textbf{Sufficient interactions lead to improved prompt adherence.}
Attention within each transformer block facilitates the integration of text and image features.
As shown in Fig.~\ref{fig:attn_map_example} (second row), different blocks exhibit different patterns, where the 7th block focuses more on structure while the 53rd block emphasizes the ``birthday'' token.
Therefore, we should consider a block-specific architecture to adaptively modulate interactions to leverage the unique capabilities of each block.
Moreover, the denoising process in DiTs is indeed a PF-ODE solving procedure~\cite{RectifiedFlow_liuflow}, where different timesteps contribute different guidance strengths from text features.
Actually, Fig.~\ref{fig:alpha}(d) shows that attention norms differ across timesteps and different blocks.
Thus, timestep-aware modulation is also necessary to manage the interaction strength dynamically.
To mitigate the problem of imbalanced visual and textual tokens ($N_{z} \gg N_{c}$) \cite{taca_lv2025rethinking}, we introduce a token-level modulation mechanism for text features with different lengths.
All these observations motivate our designs.

\begin{tcolorbox}
    [colback=gray!10, colframe=gray!70, coltext=black, title=Principle 2: Per-token interaction across transformer blocks and denoising timesteps, width=\linewidth, fontupper=\small]
    \vspace{-5pt}
    Modulating text features at block-specific, timestep-aware, and token-level dimensions is essential and effective for enhancing text-image interactions.
    \vspace{-5pt}
\end{tcolorbox}

\noindent \textbf{Differences between our \method and existing approaches.}
FiLM and adaLN provide general feature conditioning~\cite{film_perez2018,dit_AdaLN_peebles2023scalable}, while \method instead predicts a signed scalar for each text token, block, and denoising step, rescaling text states immediately before MMDiT joint attention. This changes the text keys and values seen by image queries while keeping the backbone weights frozen. LoRA~\cite{lora_hu2022} updates effective weights, whereas our learned feature modulation can also be combined with LoRA.
Unlike TACA's prescribed, timestep-aware attention amplification~\cite{taca_lv2025rethinking}, \method learns which text signals to enhance or suppress. CFG instead adjusts sampling predictions~\cite{cfg_ho2022classifier}, while \method acts inside the denoiser and improves GenEval at all seven tested guidance scales.
\begin{tcolorbox}
    [colback=gray!10, colframe=gray!70, coltext=black, title=Principle 3: Selective enhancement and suppression of text signals, width=\linewidth, fontupper=\small]
    \vspace{-5pt}
    Not all text signals contribute equally to text-image alignment. Adaptively enhancing informative signals while suppressing irrelevant ones enables effective control of cross-modal interactions without modifying the pretrained backbone.
    \vspace{-5pt}
\end{tcolorbox}

\vspace{-10pt}
\subsection{The Proposed \method}
\label{sec:our_designs}

\begin{wrapfigure}{r}{0.55\textwidth}
    \centering
    \setlength{\belowcaptionskip}{-3pt}
    \includegraphics[width=\linewidth]{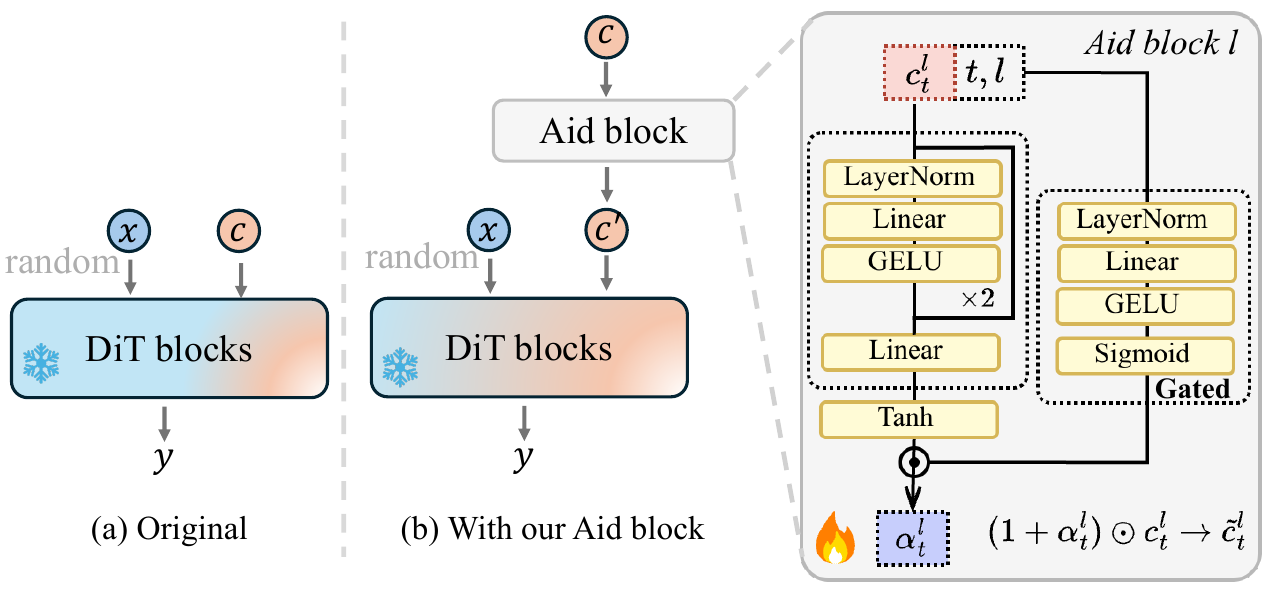}
    \caption{
    \textbf{Overview of our proposed \module blocks.}
    (a) The original diffusion transformers map a random latent and input textual conditions to output features.
    (b) Our \module blocks adaptively modulate textual conditions with respect to denoising timesteps across different transformer blocks, enhancing the interaction between image latents and textual conditions.
    }
    \label{fig:flowchart}
    \vspace{-1em}
\end{wrapfigure}

\noindent \textbf{Adaptive interaction denoising (\module) module.}
We propose \module $\phi$ that learns the coefficient $\alpha_{t}^{l} \in \mathbb{R}^{N}$ to modulate text-image interactions based on the current diffusion timestep, transformer block, and text features.
\setlength{\abovedisplayskip}{3pt}
\setlength{\belowdisplayskip}{3pt}
\setlength{\abovedisplayshortskip}{2pt}
\setlength{\belowdisplayshortskip}{2pt}
\begin{equation}
    \label{eq:alpha}
    \alpha_{t}^{l} = \phi(\mathbf{c}_t^l, t, l), \quad \alpha_{t}^{l} \in [-1, 1]^{N},
\end{equation}
where $\mathbf{c}_t^l$ represents the text features at timestep $t$ and transformer block $l$, $N$ is the number of text tokens.
Then, we modulate the text features as:
\setlength{\abovedisplayskip}{3pt}
\setlength{\belowdisplayskip}{3pt}
\setlength{\abovedisplayshortskip}{2pt}
\setlength{\belowdisplayshortskip}{2pt}
\begin{equation}
    \label{eq:modulated_text}
    \widetilde{\mathbf{c}}_t^l = \mathbf{c}_t^l + \mathbf{c}_t^l \odot \alpha_{t}^{l},
\end{equation}
where $\odot$ is element-wise multiplication.
The modulated $\widetilde{\mathbf{c}}_t^l$ enters joint attention in block $l$: positive $\alpha_t^l$ strengthens a text-token signal, negative values suppress it, and zero preserves the original state. The image-token updates are affected through the resulting text keys and values; the backbone projections remain frozen.
The \module $\phi$ is implemented as a lightweight multi-layer perceptron with a gated mechanism and bounded activation functions (\textit{i.e.}, $\tanh$) to ensure stability and effective modulation, as shown in Fig.~\ref{fig:flowchart}.

\noindent \textbf{Improving distribution sparsity across blocks.}
We expect the learned $\alpha_{t}^{l}$ to be sparse for two main reasons.
\textit{1)} Not all text tokens are equally important; some tokens carry more semantic information than others.
For example, max sequence lengths of text tokens are $512$ in FLUX and $333$ in SD 3.5, but most prompts contain only approximately $50\sim 200$ tokens, and the rest are often less informative, such as ``PAD'' tokens.
\textit{2)} Not all transformer blocks require strong text conditioning.
Commonsense from previous works~\cite{alaluf2024cross,li2026unraveling} indicates that early blocks focus more on semantic and structural information, while later blocks refine visual details.
Therefore, the $\alpha_{t}^{l}$ should be sparse, allowing the model to focus on the most relevant tokens and transformer blocks.

To achieve this, we facilitate two strategies to enable that the \module $\phi$ learns an implicit and efficient $\alpha$ for block-specific and token-level modulation.
Inspired by Qiu et al.~\cite{gated_attention_qiu2025}, we apply a gated mechanism in \module to promote sparsity:
\setlength{\abovedisplayskip}{3pt}
\setlength{\belowdisplayskip}{3pt}
\setlength{\abovedisplayshortskip}{2pt}
\setlength{\belowdisplayshortskip}{2pt}
\begin{equation}
    \label{eq:gated_alpha}
    \alpha_{t}^{l} = \phi(\mathbf{c}_t^l, t, l)= \tanh(\mathbf{c}_t^lW_l) \cdot \sigma(\mathbf{c}_t^lW_g),
\end{equation}
where $W_l$ represents the feature weights, and $W_g$ represents the gating weights. The gate score $\sigma(\mathbf{c}_t^lW_g)$ effectively acts as a dynamic filter, controlling the strength of the text features by selectively allowing or blocking the modulation signal.
Furthermore, we introduce a regularization term on the modulation coefficients, which penalizes large values of $\alpha_{t}^{l}$:
\setlength{\abovedisplayskip}{3pt}
\setlength{\belowdisplayskip}{3pt}
\setlength{\abovedisplayshortskip}{2pt}
\setlength{\belowdisplayshortskip}{2pt}
\begin{equation}
    \label{eq:reg_loss}
    \mathcal{L}_{reg} = \mathbb{E}_{t,l} \left[ \|\alpha_{t}^{l}\|_2 \right].
\end{equation}

\vspace{-10pt}
\subsection{Training and Inference}
\module modules are trained using the standard diffusion loss, which minimizes the squared error between the predicted velocity and the target velocity ($\mathbf{\epsilon} - \mathbf{x}$):
{
\setlength{\abovedisplayskip}{3pt}
\setlength{\belowdisplayskip}{3pt}
\setlength{\abovedisplayshortskip}{2pt}
\setlength{\belowdisplayshortskip}{2pt}
\begin{equation}
    \label{eq:diffusion_loss}
    \mathcal{L}_{diff} = \mathbb{E}_{t, \mathbf{\epsilon},\mathbf{x}} \left[ \| v_\theta(\mathbf{z}_t, t, \widetilde{\mathbf{c}}_t) - (\mathbf{\epsilon} - \mathbf{x}) \|_2^2 \right],
\end{equation}
}
where $v_\theta$ is the velocity predicted by the diffusion model.
During training, we freeze the backbone MMDiT model and only optimize the \module $\phi$ using the combined loss:
{
\setlength{\abovedisplayskip}{3pt}
\setlength{\belowdisplayskip}{3pt}
\setlength{\abovedisplayshortskip}{2pt}
\setlength{\belowdisplayshortskip}{2pt}
\begin{equation}
    \label{eq:total_loss}
    \mathcal{L} = \mathcal{L}_{diff} + \lambda_{reg} \mathcal{L}_{reg},
\end{equation}
}
where $\lambda_{reg}$ is the regularization hyperparameter.
To make the optimization more stable and $\alpha_{t}^{l}$ more robust, we randomly skip the \module modules during training with a probability of $p$.
This mutes the entire \module rather than individual neurons, akin to dropout.

During inference, the lightweight \module modules are all active and inserted into each block.
Other than the integration of our proposed \module, the inference process remains unchanged. The module is trained beforehand; at inference it recomputes $\alpha_t^l$ from the evolving text states at each denoising step and block.
Regarding downstream applications such as LoRAs and editing, our weights are directly loaded for zero-shot inference.

\vspace{-5pt}
\section{Experiments}
\vspace{-5pt}
\label{sec:experiments}

\begin{figure}[t]
    \centering
    \includegraphics[width=\linewidth]{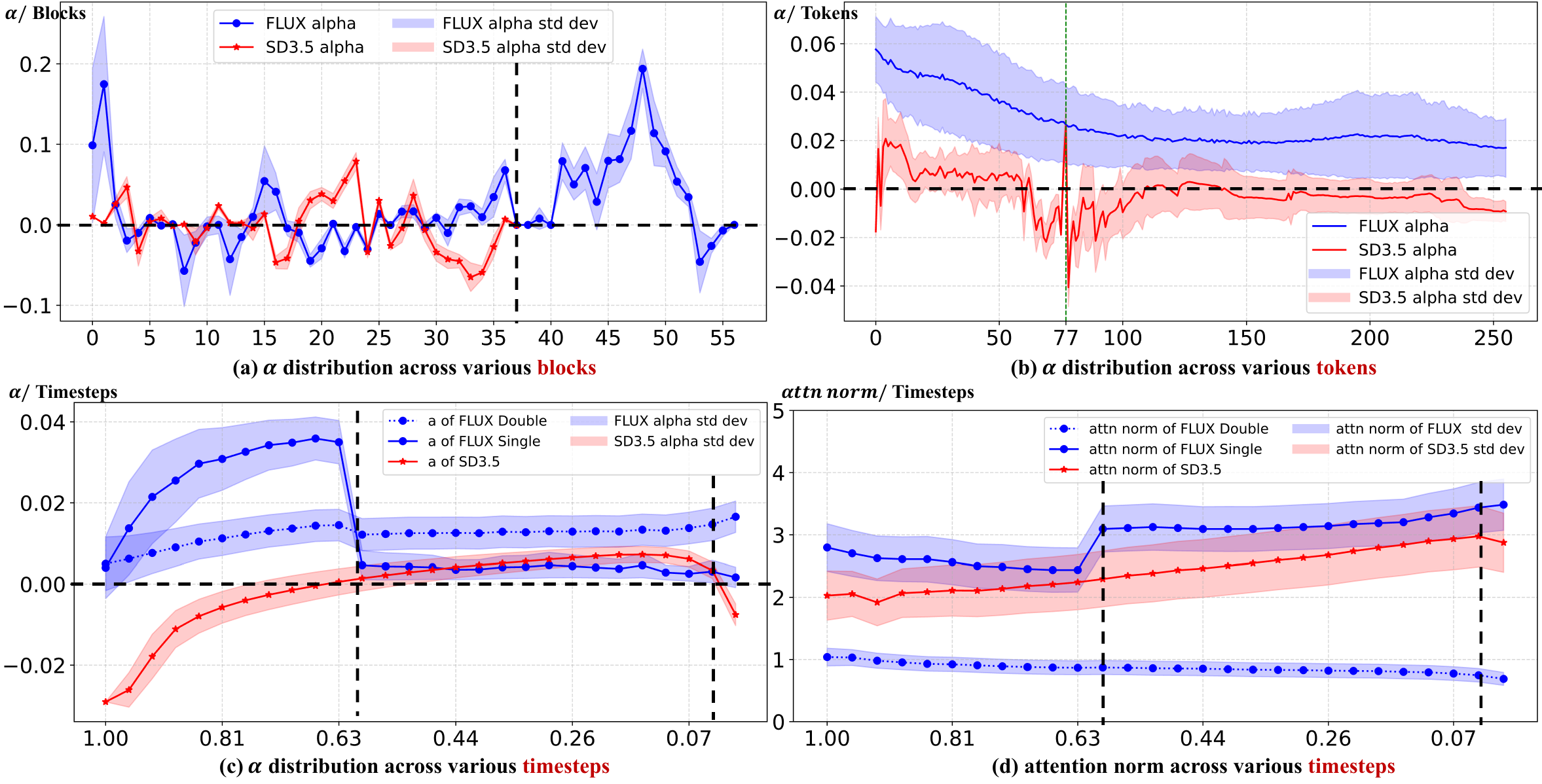}
    \vspace{-0.5em}
    \caption{
    \textbf{Distributional visualization of the learned adaptive $\alpha$ across various (a) diffusion transformer blocks , (b) textual tokens and (c) denoising timesteps.}
    (d) Visualization of the attention norm across various timesteps.
    For (c) and (d), we separately show FLUX single-stream and dual-stream results to highlight architectural differences.
    Curves show means across $2,400$ held-out prompts while shaded bands show prompt-level standard deviations after averaging over the other displayed axes.
    Positive and negative values indicate enhancement and suppression rather than a universal trajectory for every prompt.
    }
    \label{fig:alpha}
    \vspace{-0.5em}
\end{figure}

\noindent \textbf{Implementation details.}
We adopt FLUX.1-Dev~\cite{flux_2024} and SD3.5-Large~\cite{SD35_2024} as baselines.
Each \module module consists of a two-layer MLP with hidden dimension $d=2048$ and a gated mechanism.
During training, we freeze all backbone parameters and only optimize the \module modules, introducing only a 1.51\% parameter increase over the baseline FLUX.1-Dev.
For optimization, we use the Prodigy optimizer~\cite{prodigy_mishchenko2024} with a learning rate of $1.0$ and gradient accumulation of $8$, yielding an effective batch size of $64$.
Training runs for $4{,}000$ steps ($3.3$ epochs) on $8$ H100 GPUs(approximately $6$ hours for FLUX and $4.5$ hours for SD 3.5).
We set the dropout probability $p=0.1$, and $\lambda_{reg}=0.1$.
During inference, all the \module modules are activated.
We sample $1024 \times 1024$ images using the EDM~\cite{karras2022elucidating} sampler for $28$ steps, with CFG scales of $7.0$ (SD3.5) and $3.5$ (FLUX).

\noindent \textbf{Datasets and evaluation metrics.}
We train on a subset of HPDv3~\cite{HPSV3_ma2025hpsv3} comprising $77,252$ text-image pairs, filtered by preference confidence and image source (Appendix~\ref{appendix:data_quality}).
We evaluate \method on the full HPDv3 benchmark for image quality and human preference alignment, as well as GenEval~\cite{geneval_ghosh2023} for semantic understanding.
In addition, we assess the generalizability of \method using out-of-distribution metrics such as HPSv2~\cite{HPSV2_wu2023human}, ImageReward~\cite{xu2023imagereward} and Aesthetic Score~\cite{laion_aesthetic_2022}.
In Sec.~\ref{sec:applications}, we demonstrate \method's compatibility with domain-specific LoRA and ControlNet modules using subsets of HPDv3, and evaluate its performance on editing tasks with EditBench~\cite{editbench_lin2024schedule}.

\begin{figure*}[t]
    \centering
    \includegraphics[width=\linewidth]{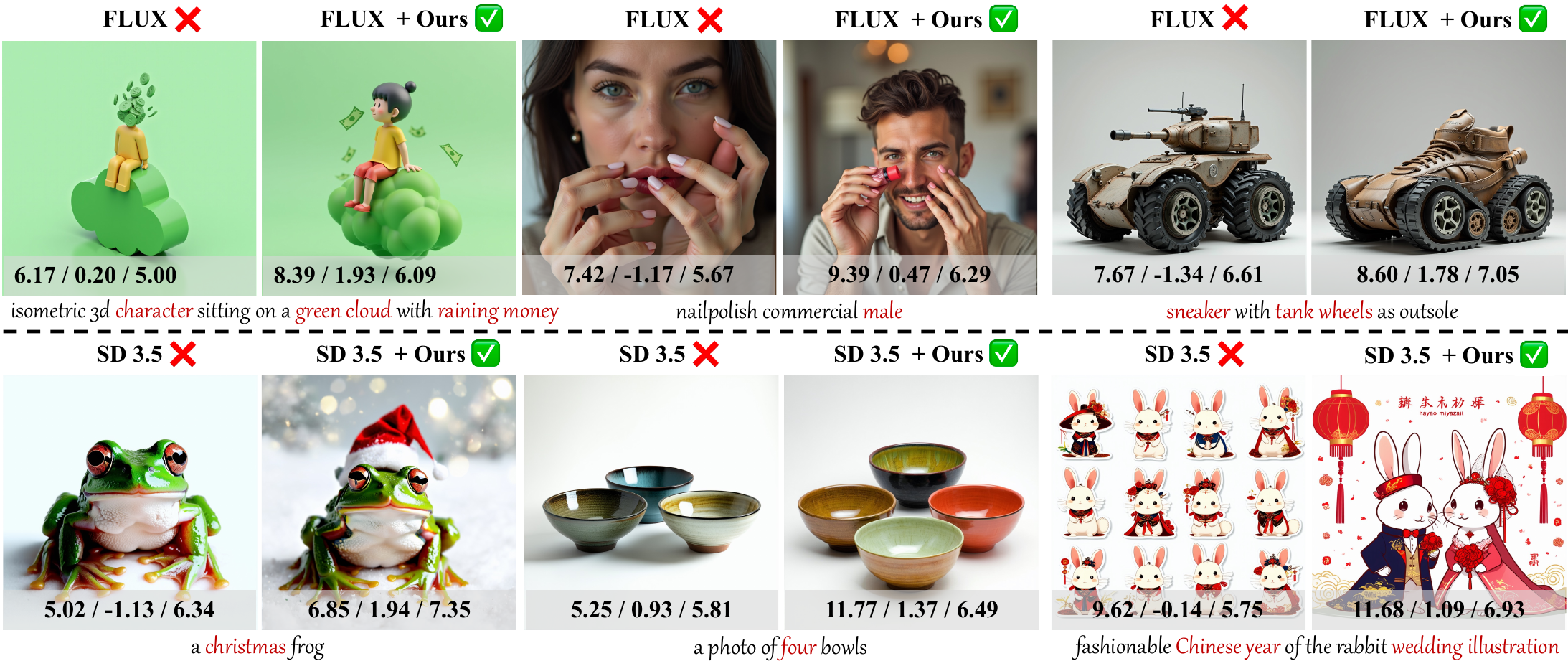}
    \vspace{-0.5em}
    \caption{
    \textbf{Qualitative comparisons of images generated by baseline FLUX and SD 3.5 with and without our proposed \method.}
    The quantitative results below each image shows the scores of HPSv3, Image Reward and Aesthetic.
    Our method demonstrates superior prompt adherence and image quality.
    More qualitative results are given in appendix.
    }
    \label{fig:main_showcase}
    \vspace{-1.0em}
\end{figure*}

\subsection{Empirical Observation of Learned $\alpha$}

A key contribution of our work is that the learned $\alpha$ exhibits strong interpretability, highly correlating with the model's internal interaction patterns.
To verify that our \method learns to adaptively adjust block-wise interactions between token-level textual conditions across denoising trajectories, we analyze the learned coefficients $\alpha_t^l \in \mathbb{R}^{N}$ cached in each \module block during denoising.
Fig.~\ref{fig:alpha} summarizes the learned $\alpha$ and attention norms across the same 2,400-prompt evaluation subset.
The prompt-specific token patterns are visualized in Appendix~\ref{appendix:adaptive_modulation_visualization}.

\noindent \textbf{$\alpha$ adaptively exhibits sparse and independent patterns across different transformer blocks.}
The mean and standard deviation of $alpha$ over dimensions $\{T,N\}$ for each block are shown in Fig.~\ref{fig:alpha}(a).
We could observe that the former blocks ($1st-3rd$ in FLUX and $2nd-4th$ in SD 3.5) and the latter blocks ($41st-52nd$ in FLUX and $30th-35th$ in SD 3.5) exhibit larger $\alpha$ values than the middle blocks.
This aligns well with the observations~\cite{hertz2022prompt,cao2023masactrl,li2026unraveling} that early blocks focus on coarse structure and late blocks refine fine details.
Additionally, there are noticeable zero regions in several blocks, showing the effectiveness of our sparsity strategies.
The signed $\alpha$ learned by \method captures varying block contributions, including suppression in selected blocks. In a separate lightweight ($d=512$) control, clamping negative coefficients at inference reduces HPSv3 from $10.82$ to $10.72$ and GenEval from $0.709$ to $0.692$ (Appendix~\ref{appendix:sparse_enhance_results}).

\noindent \textbf{$\alpha$ assigns token-specific importance weights to different textual tokens.}
The $\alpha$ distribution across token positions is shown in Fig.~\ref{fig:alpha}(b).
The vertical line indicates the boundary between two concatenated text encoders in SD 3.5.
In total, $\alpha$ shows a declining trend as token position increases for both models.
This is expected since earlier tokens typically carry more semantic weight in prompts, while later tokens often include less informative padding or filler words.
Notably, the abrupt change around token index $77$ in SD 3.5 coincides with the two text encoders' boundary.
All these observations confirm that our \method accurately captures token-level importance variations.

\noindent \textbf{$\alpha$ implicitly learns dynamic timestep awareness throughout denoising trajectories.}
To investigate the timestep-awareness of $\alpha$, we visualize its distribution throughout the denoising process in Fig.~\ref{fig:alpha}(c).
Each block's $\alpha$ values are shown in Fig.~\ref{fig:alpha_timestep_heatmap_sd35} and Fig.~\ref{fig:alpha_timestep_heatmap_flux} of appendix.
FLUX is split into two parts due to its single-stream and dual-stream architectures.
The different types of blocks behave differently, while single-stream blocks show a clear drop in $\alpha$ values at $t=0.607$, while the dual-stream blocks and SD 3.5 blocks exhibit smooth changes.
Such trend indicates that \method implicitly learns to adjust text-image interactions dynamically across timesteps.

\noindent \textbf{$\alpha$ aligns the intrinsic attributes of pretrained models for efficient and effective adaptation.}
To explain the $\alpha$ changes across denoising trajectories, we visualize the attention norm of textual and image features in attention blocks across timesteps in Fig.~\ref{fig:alpha}(d).
We could clearly observe that the dashed vertical lines in Fig.~\ref{fig:alpha}(d) perfectly align with the sharp changes in $\alpha$ in Fig.~\ref{fig:alpha}(c).
Based on the observations, our learned $\alpha$ already captures the intrinsic attribute changes of pretrained models across timesteps, enabling efficient and effective adaptation for the text-image interactions.

\begin{table}[!t]
    \centering
    \caption{%
    \textbf{
    Quantitative comparisons of strong baselines and our proposed \method on the HPSv3 benchmark.
    }
    \method significantly improves the baseline models across categories and even outperforms the current leading Kolors in most categories when applied to FLUX.
    $^\dagger$ denotes results evaluated by ourselves and $^*$scores are quoted from the official HPSv3 leaderboard~\cite{HPSV3_ma2025hpsv3}.
    \textbf{Bold} and \underline{underline} indicate the best and second-best results, respectively.
    }
    \label{tab:hpsv3_results}
    \small
    \setlength{\tabcolsep}{0.18mm}
    \resizebox{\linewidth}{!}{%
        \begin{tabular}{lcccccccccccccc}
        \toprule
        Models & All & Char. & Arts & Design & Archi. & Animals & Scenery & Transpor.& Products & Plants & Food & Science & Others \\
        \midrule
        Kolors~\cite{kolors_2024}$^*$ & \underline{10.55} & \underline{11.79} & \underline{10.47} & \underline{9.87} & 10.82 & \underline{10.60} & 9.89 & 10.68 & \underline{10.93} & \underline{10.50} & 10.63 & \underline{11.06} & 9.51 \\
        FLUX~\cite{flux_2024}$^*$ & 10.43 & 11.70 & 10.32 & 9.39 & 10.93 & 10.38 & 10.01 & 10.84 & \textbf{11.24} & 10.21 & 10.38 & \textbf{11.24} & 9.16 \\
        Infinity~\cite{han2025infinity}$^*$ & 10.26 & 11.17 & 9.95 & 9.43 & 10.36 & 9.27 & \underline{10.11} & 10.36 & 10.59 & 10.08 & 10.30 & 10.59 & 9.62 \\
        SD 3~\cite{SD3_esser2024scaling}$^*$ & 5.31 & 6.70 & 5.98 & 5.15 & 5.25 & 4.09 & 5.24 & 4.25 & 5.71 & 5.84 & 6.01 & 5.71 & 4.58 \\
        \midrule
        SD 3.5$^\dagger$ & 9.31 & 9.66 & 8.63 & 8.58 & 10.03 & 8.81 & 8.90 & 9.58 & 9.67 & 9.78 & 10.05 & 8.60 & 9.46 \\
        SD 3.5 + TACA$^\dagger$ & 9.44 & 9.81 & 8.82 & 8.54 & 10.31 & 8.92 & 9.26 & 9.54 & 9.96 & 9.85 & 10.05 & 8.81 & 9.48 \\
        SD 3.5 + Ours & 9.48 & 9.93 & 8.87 & 8.69 & 10.22 & 9.00 & 9.17 & 9.80 & 9.76 & 9.87 & 10.18 & 8.74 & 9.55\\
        \midrule
        FLUX$^\dagger$ & 10.42 & 11.67 & 10.31 & 9.35 & 10.94 & 10.37 & 9.97 & 10.87 & 10.48 & 10.39 & \textbf{11.20} & 9.24 & 10.24 \\
        FLUX + TACA$^\dagger$ & 10.48 & 11.72 & 10.35 & 9.36 & \underline{10.97} & 10.45 & 10.03 & \underline{11.04} & 10.53 & 10.46 & \textbf{11.20} & 9.33 & \underline{10.35} \\
        FLUX + Ours & \textbf{10.71} & \textbf{11.96} & \textbf{10.72} & \textbf{9.90} & \textbf{11.19} & \textbf{10.62} & \textbf{10.18} & \textbf{11.24} & 10.61 & \textbf{10.67} & \underline{11.18} & 9.68 & \textbf{10.58} \\
        \bottomrule
        \end{tabular}
    }
\end{table}

\subsection{Main Results}

\noindent \textbf{Qualitative results.}
To demonstrate the effectiveness of our proposed \method, we present qualitative comparisons of images generated by different models given the same prompts and seeds in Fig.~\ref{fig:teaser} and Fig.~\ref{fig:main_showcase}.
FLUX with \method accurately captures the ``\textit{Nailpolish}'' and ``\textit{Male}'', generates a correct quantity of ``\textit{donuts}'' and ``\textit{sneaker with tank wheels}''.
SD 3.5 with \method effectively synthesizes the ``\textit{christmas frog}'' and ``\textit{rabbit wedding}'' with semantic details and better visual quality.
Although derived from the same noise, \method adaptively denoises to better trajectories according to the prompts, demonstrating its effectiveness in enhancing prompt adherence and image quality.

\noindent \textbf{Quantitative results on HPSv3 benchmark.}
We conduct quantitative comparisons on the HPSv3 benchmark~\cite{HPSV3_ma2025hpsv3}.
The results are summarized in Tab.~\ref{tab:hpsv3_results}.
Both FLUX and SD 3.5 benefit from our \method, with improvements of $0.29$ and $0.17$ in overall HPSv3 score, compared to their respective baselines.
Our method also competes favorably against TACA~\cite{taca_lv2025rethinking}, and FLUX combined with \method even outperforms current SoTA Kolors~\cite{kolors_2024} in overall score and most categories.
These significant improvements across all categories validate \method's capability towards diverse and complex prompts, leading to improved human preferences.

\noindent \textbf{Computational overhead analysis.}
To evaluate the efficiency of our \method, we measure the computational overhead on $1,000$ images under default settings.
With hidden dimension $d=2048$ of the \module, \method introduces only $179.77M$ additional parameters ($1.51\%$ over the FLUX), and increases inference time by only $0.08$ seconds (from $5.75s$ to $5.83s$).
\method achieves substantial performance improvements with negligible computational cost. The separate $d=512$ variant adds $0.38\%$ parameters.
Details are provided in Appendix~\ref{appendix:parameter_efficiency}.

\noindent \textbf{Generalization to other metrics and benchmark.}
Although \method is optimized solely on the HPDv3 dataset, it brings consistent improvements on widely-used metrics such as HPSv2, ImageReward, and Aesthetic Score.
The results are summarized in Tab.~\ref{tab:more_metric_results}.
HPSv2 scores improve by $0.15$ and $0.20$ for SD 3.5 and FLUX when combined with \method, respectively.
Moreover, \method also improves the GenEval performance, achieving $5\%$ and $2\%$ improvements for SD 3.5 and FLUX, respectively.
These results validate the effectiveness and generalizability of our approach across different evaluation metrics and benchmarks.
More results  are in Appendices~\ref{appendix:alignment_benchmarks},~\ref{appendix:additional_baselines} and ~\ref{appendix:cfg_sweep}.

\noindent \textbf{Human evaluation.}
We conduct blinded pairwise comparisons against FLUX, SD~3.5, and FLUX+LoRA.
For each comparison, we select $200$ prompts and generate two image pairs per prompt, matching the prompt and random seed between the methods in each pair.
Eight volunteers judge each pair by choosing which image is better or indicating that they are equally good.
As is shown in Tab.~\ref{tab:human_evaluation},
\method receives more votes than the alternative in all three comparisons: $44.50\%$ versus $15.84\%$ against FLUX, $41.81\%$ versus $22.63\%$ against SD~3.5, and $41.00\%$ versus $26.50\%$ against FLUX+LoRA.
Our method achieved a consistent advantage in human evaluation experiments.

\begin{table}[tb]
    \small
    \centering
    \caption{\textbf{Blinded pairwise human preferences.} Each comparison uses 200 prompts, two generated image pairs per prompt, and eight volunteer judgments per pair ($3{,}200$ votes).}
    \label{tab:human_evaluation}
    \setlength{\tabcolsep}{3mm}{
    \begin{tabular}{lrrr}
        \toprule
        Pair & Alternative wins & Tie & \method wins \\
        \midrule
        FLUX vs. FLUX + \method & 507 (15.84\%) & 1,269 (39.66\%) & 1,424 (44.50\%) \\
        SD~3.5 vs. SD~3.5 + \method & 724 (22.63\%) & 1,138 (35.56\%) & 1,338 (41.81\%) \\
        FLUX+LoRA vs. FLUX + \method & 848 (26.50\%) & 1,040 (32.50\%) & 1,312 (41.00\%) \\
        \bottomrule
    \end{tabular}
    }
    \vspace{-1em}
\end{table}

\subsection{Ablation Studies}
\label{sec:ablations}
To identify the efficacy of each component and optimization choice in our \method, here we perform ablation studies by disabling each component for evaluation.
Specifically, we train various variants on the FLUX baseline following the same settings in Sec.~\ref{sec:experiments} except specific component is removed.
We calculate quantitative scores on a random $20\%$ subset of HPDv3 benchmark for a more efficient investigation.
The quantitative results are presented in Tab.~\ref{tab:ablation_study}.

\noindent \textbf{Ablation on boundedness and sparsity constraints.}
Recall that we introduce a regularization term and a gated mechanism to respectively stabilize and sparsify the values of $\alpha$ in Sec.~\ref{sec:methods}.
The results in Tab.~\ref{tab:ablation_study}~(\textit{``Bounded Constraints''} and \textit{``Sparsity Constraints''} groups) show that the performance degrades when these constraints are removed.
Compared to simply clipping the $\alpha$ values, using Tanh activation yields better results($10.48$ vs. $7.12$ on HPSv3).
Adding a smaller regularization weight ($\lambda_{reg}=0.1$) further improves the performance than a larger one ($\lambda_{reg}=1.0$).
Armed with the gated mechanism, the model achieves the best results ($28.71$ on HPSv2, $10.74$ on HPSv3) compared to the preceding settings.
Ablation studies validate the effectiveness of our proposed boundedness and sparsity constraints in stabilizing and enhancing the learning of $\alpha$.

\noindent \textbf{Ablation on optimization choices and \module drop probability.}
We investigate the impact of different optimization strategies and dropout probabilities $p$ during training.
As shown in Tab.~\ref{tab:ablation_study}~(\textit{``Optimization Choices''} group), LoRA-based methods yield only marginal improvements over the FLUX baseline.
In contrast, all variants of our \module modules substantially outperform LoRA-based approaches across all metrics.
Notably, our \module modules can be effectively integrated with reward model or DPO to achieve competitive results.
For optimization simplicity and to highlight the $\alpha$ learning mechanism, we adopt the straightforward SFT as our default optimization strategy.
Regarding the dropout probability $p$ in Tab.~\ref{tab:ablation_study}~(\textit{``Drop p''} group), we observe that a small $p=0.1$ consistently achieves the best performance across all metrics, outperforming both $p=0.3$ and $p=0.5$.
These results validate that our \module design with adaptive $\alpha$ learning is inherently more effective than LoRA-based methods.

\noindent \textbf{Ablation of adaptive dimensions.}
To quantify the individual contribution of each dimension in our design, we remove the adaptivity of one dimension while keeping the other two fully adaptive.
Tab.~\ref{tab:ablation_study}~(\textit{``Dimension Adaptivity''} group) isolates the block, text-token, and timestep dimensions on the same 2,400-prompt HPDv3 subset.
Sharing the module across blocks or a coefficient across text tokens lowers HPSv3 from $10.84$ to $9.58$ or $10.28$, respectively.
Time-averaging the coefficients lowers it to $10.46$, while fixing them to one denoising step performs worse.
These results show the necessity of adaptive modulation in our \method.
Appendix~\ref{appendix:dimension_ablation} specifies the details.

\begin{table}[t]
\footnotesize
\centering
\begin{minipage}[t]{0.48\textwidth}
    \centering
    \caption{
    \textbf{Quantitative comparisons of more image quality metrics.}
    Despite trained only on HPDv3 dataset, our \method generalizes to other metrics and gains consistent improvements on these metrics.
    Note that GenEval scores are computed from official GenEval benchmarks.
    }
    \label{tab:more_metric_results}
    \setlength{\tabcolsep}{0.6mm}{
    \begin{tabular}{lccccc}
        \toprule
        Models & HPSv2 & HPSv3 & ImgRwd & Aes. & GenEval  \\ \midrule
        SD 3.5 & \underline{28.86} & 9.31 & \underline{0.94} & 6.42 & 0.72 \\
        SD 3.5 + TACA & 28.89  &  \underline{9.44} & 0.93 & \underline{6.44} & \underline{0.73} \\
        SD 3.5 + Ours & \textbf{29.01} & \textbf{9.48} & \textbf{1.05} & \textbf{6.53} & \textbf{0.77} \\
        \midrule
        FLUX & 28.53 & 10.42 & 0.89 & 6.65 & \underline{0.68} \\
        FLUX + TACA & 28.59 & \underline{10.48} & 0.90 & \underline{6.68} & \underline{0.68} \\
        FLUX + Ours & 28.73 & \textbf{10.71} & \underline{0.94} & \textbf{6.73} & \textbf{0.70} \\
        \bottomrule
    \end{tabular}
    }

    \caption{
    \textbf{Quantitative comparison of instructional image editing tasks on EditBench benchmark.}
    Despite trained on the pure generation settings, our \method generalizes to homologous Kontext by enhancing test-time interactions between editing prompts and images in a zero-shot manner.
    }
    \label{tab:metric_editing_task}
    \setlength{\tabcolsep}{0.2mm}{
    \resizebox{\linewidth}{!}{%
    \begin{tabular}{lcccccc}
        \toprule
        Models & HPSv2 & HPSv3 & ImgRwd & Aes. & CLIP-I & CLIP-T \\
        \midrule
        Kontext        & 27.48 & 8.50 & \textbf{0.62} & 6.51 & 90.74 & 26.35 \\
        Kontext + Ours & \textbf{27.54} & \textbf{8.62} & \textbf{0.62} & \textbf{6.53} & \textbf{91.67} & \textbf{26.41} \\
        \bottomrule
    \end{tabular}
    }
    }
\end{minipage}
\hfill
\begin{minipage}[t]{0.49\textwidth}
    \centering
    \caption{
    \textbf{Ablation study on key components of our proposed method}.
    ``+'' means adding the new component on the previous setting.
    The \textbf{\textit{italicized}} settings are our default choices.
    }
    \label{tab:ablation_study}
    \setlength{\tabcolsep}{0.6mm}{
    \begin{tabular}{llcccc}
        \toprule
        Categories & Settings & HPSv2 & HPSv3 & ImgRwd & Aes. \\
        \midrule
        FLUX & Baseline & 28.53 & 10.42 & 0.89 & 6.66 \\
        \midrule
        \multirow{3}{*}{\begin{tabular}[c]{@{}l@{}}Bounded\\ Constraints\end{tabular}} & No Const. & 20.03 & 6.41 & -2.04 & 3.54 \\
          & + Clip & 22.86 & 7.12 & 0.31 & 4.07 \\
          & \textbf{\textit{+ Tanh Act}} & 28.55 & 10.48 & 0.90 & 6.67 \\
        \midrule
        \multirow{3}{*}{\begin{tabular}[c]{@{}l@{}}Sparsity\\ Constraints\end{tabular}} & + Reg(1.0) & 28.49 & 10.44 & 0.90 & 6.64 \\
        & \textit{+ Reg(0.1)} & 28.61 & 10.59 & 0.92 & 6.68 \\
        & \textbf{\textit{+ Gated}} & 28.71 & 10.74 & 0.94 & 6.72 \\
        \midrule
        \multirow{3}{*}{\begin{tabular}[c]{@{}l@{}} Drop $p$\end{tabular}} & + drop(0.5) & 28.21 & 10.19 & 0.77 & 6.59 \\
            & + drop(0.3) & 28.47 & 10.44 & 0.87 & 6.64 \\
          & \textbf{\textit{+ drop(0.1)}} & 28.80 & 10.84 & 0.95 & 6.76 \\
        \midrule
        \multirow{5}{*}{\begin{tabular}[c]{@{}l@{}}Optimization\\ Choices\end{tabular}} & LoRA +SFT & 28.54 & 10.43 & 0.86 & 6.63 \\
            & LoRA +DPO & 28.57 & 10.48 & 0.90 & 6.67 \\
            & Ours +Rwd & 28.76 & 10.78 & 0.93 & 6.73 \\
            & Ours +DPO & 28.81 & 10.85 & 0.94 & 6.75 \\
            & \textbf{\textit{Ours +SFT}} & 28.80 & 10.84 & 0.95 & 6.76 \\
        \midrule
        \multirow{3}{*}{\begin{tabular}[c]{@{}l@{}}Dimension\\ Adaptivity \end{tabular}} & \textit{w/o} blocks & 27.41 & 9.58 & 0.77 & 6.34 \\
         & \textit{w/o} tokens & 28.07 & 10.28 & 0.88 & 6.48 \\
         & \textit{w/o} timesteps & 28.53 & 10.32 & 0.82 & 6.35 \\ \midrule
         FLUX + Ours & \multicolumn{1}{c}{\textbf{\textit{full}}} & 28.80 & 10.84 & 0.95 & 6.76 \\
        \bottomrule
    \end{tabular}
    }
\end{minipage}
\end{table}

\vspace{-10pt}
\subsection{Generalization to Downstream Applications}
\label{sec:applications}
\vspace{-4pt}

\begin{figure*}[t]
    \centering
    \begin{subfigure}[t]{0.31\linewidth}
        \centering
        \setlength{\abovecaptionskip}{-5pt}
        \setlength{\belowcaptionskip}{0pt}
        \includegraphics[width=\linewidth]{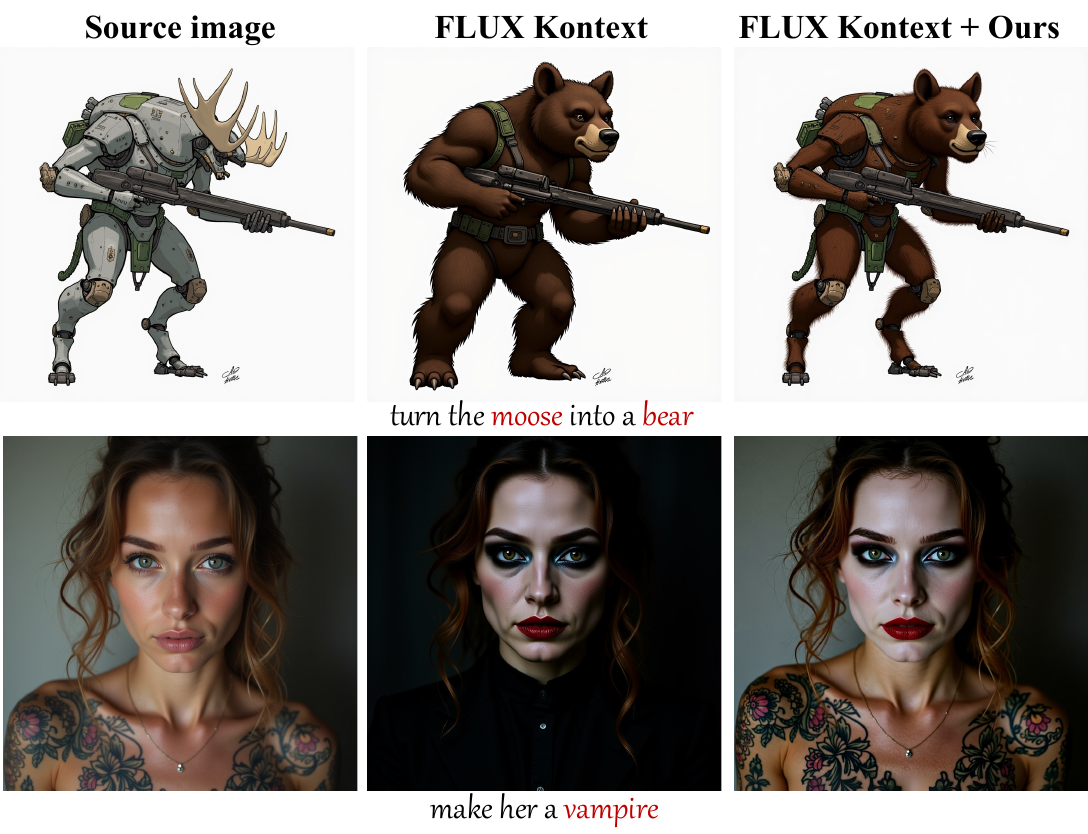}
        \vspace{-0.5em}
        \caption{
        \textbf{Instructional image editing.}
        }
        \label{fig:editing_qualitative}
    \end{subfigure}
    \hfill
    \begin{subfigure}[t]{0.31\linewidth}
        \centering
        \setlength{\abovecaptionskip}{-5pt}
        \setlength{\belowcaptionskip}{0pt}
        \includegraphics[width=\linewidth]{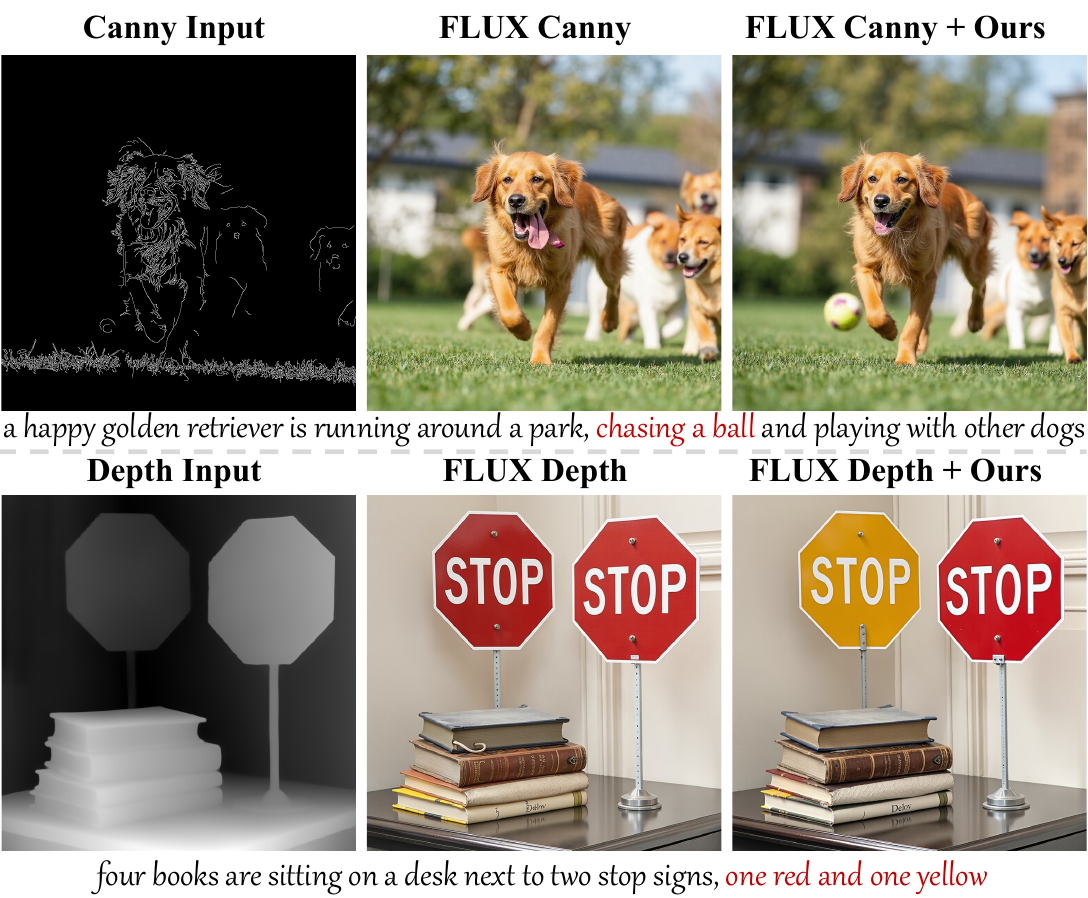}
        \vspace{-0.5em}
        \caption{
        \textbf{Controllable generation.}
        }
        \label{fig:ctrl_qualitative}
    \end{subfigure}
    \hfill
    \begin{subfigure}[t]{0.31\linewidth}
        \centering
        \setlength{\abovecaptionskip}{0pt}
        \setlength{\belowcaptionskip}{0pt}
        \includegraphics[width=\linewidth]{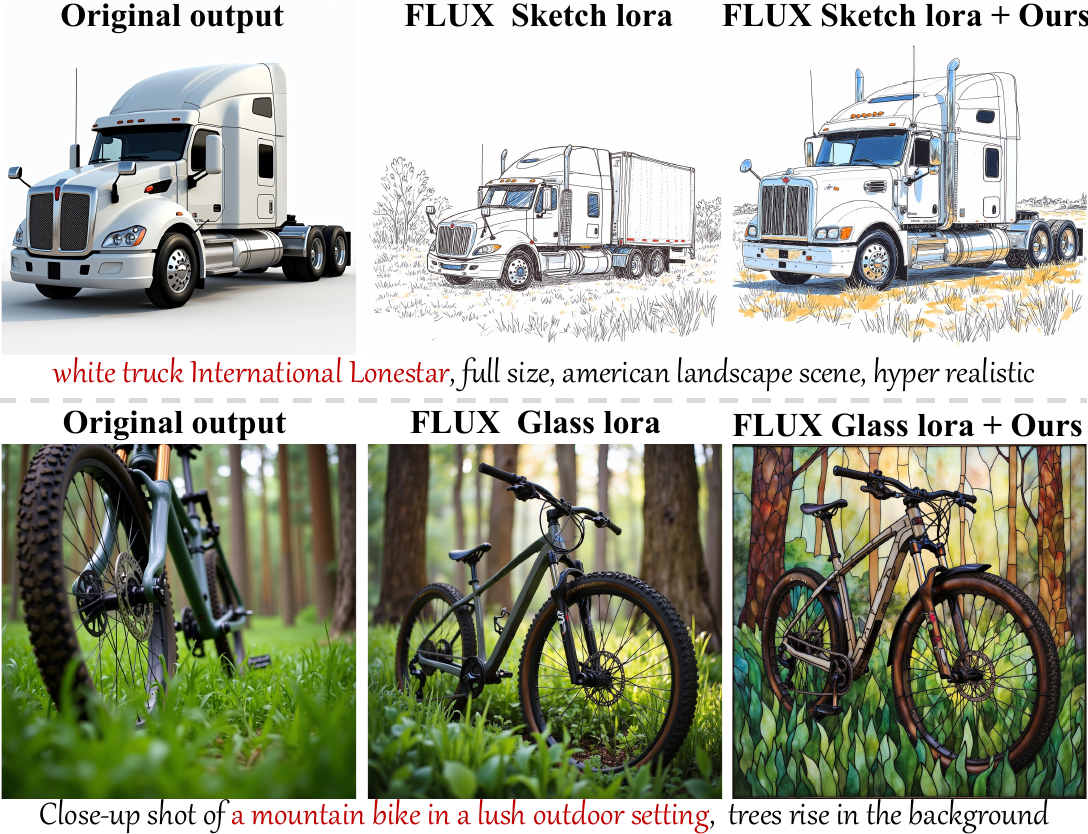}
        \caption{
        \textbf{LoRA combination.}
        }
        \label{fig:lora_qualitative}
    \end{subfigure}
    \caption{
    \textbf{Qualitative results of \method on downstream applications.} \method demonstrates strong generalization across diverse tasks without task-specific training, improving instruction-based editing, precise controllable generation, and high-quality stylized synthesis.
    }
    \label{fig:downstream_apps}
    \vspace{-1em}
\end{figure*}

\noindent \textbf{Generalizing to zero-shot instruction-based editing.}
Our \module blocks can be directly integrated into FLUX.1 Kontext~\cite{flux_kontext_labs2025flux} without any additional training.
We evaluate the edited images on the EditBench benchmark~\cite{editbench_lin2024schedule}, as summarized in Tab.~\ref{tab:metric_editing_task}.
All the metrics show consistent improvements when \method is applied to Kontext.
Notably, CLIP-I and CLIP-T scores increase by $0.93$ and $0.06$, respectively, indicating better alignment with editing instructions.
Visualizations in Fig.\ref{fig:teaser} and Fig.~\ref{fig:editing_qualitative} further keep the original details while accurately following editing instructions.

\noindent \textbf{Integrating with conditional control models.}
Cooperating with ControlNet~\cite{controlnet_zhang2023}, our \method can further enhance controllable generation by improving the interactions between complex textual conditions.
In Fig.~\ref{fig:teaser} and Fig.~\ref{fig:ctrl_qualitative}, our method effectively captures the semantic details, such as the ``\textit{Wolf}'' and  ``\textit{chasing a ball}'', which are missed by the baseline.
Moreover, ``\textit{one yellow}'' stop sign is accurately generated with \method, showing its effectiveness in handling complex textual conditions.

\noindent \textbf{Combining with existing LoRAs.}
When combined with off-the-shelf LoRAs from the community, our \method boosts the generation quality while preserving the desired styles.
In Fig.~\ref{fig:teaser} and Fig.~\ref{fig:lora_qualitative}, the images generated by combining \method exhibit richer details and more vivid textures.
The quantitative results are shown in Tab.~\ref{tab:appendix_application_results} in Appendix~\ref{sec:appendix_applications}.

\vspace{-4pt}
\section{Related Works}
\vspace{-4pt}
\label{sec:related_works}

\noindent \textbf{Diffusion transformers.}
DiT~\cite{dit_AdaLN_peebles2023scalable} pioneers the use of transformers~\cite{attention_vaswani2017} in diffusion models for image and video generation.
Following DiT, Multimodal Diffusion Transformers (MMDiTs)~\cite{SD3_esser2024scaling,flux_2024,qwenimage_wu2025technicalreport} facilitate text-image interactions through a unified transformer.
Subsequent works~\cite{hunyuan_image_2025,seedream2_gong2025,OpenAI2024_Sora,wan2025wan,wang2026klear,zhang2025uniavgen,li2026generating} have promoted AI generated content (AIGC).

\noindent \textbf{Improving cross-modal interactions in diffusion models.}
Early works~\cite{rombach2022high,kolors_2024} inject text features into diffusion models via cross-attention mechanisms.
DiT~\cite{dit_AdaLN_peebles2023scalable,pixart_chen2024} uses adaLN-zero to condition transformer activations, while FiLM~\cite{film_perez2018} applies feature-wise affine conditioning.
MMDiTs~\cite{SD3_esser2024scaling,flux_2024,qwenimage_wu2025technicalreport} suggest that text and image features should be treated equally by jointly modeling them in a unified transformer architecture.
Other paradigms include in-context conditioning~\cite{huang2024incontextlora} and auxiliary controlnets~\cite{controlnet_zhang2023,t2i_adapter_mou2024}.
Related studies explore instance-aware prompt adaptation~\cite{xiao2025visual}, layer-specific feature fusion~\cite{xiao2026layer}, and causal attention interventions~\cite{zhu2026craft} to enhance feature adaptation and multimodal interactions.
Our intervention instead rescales individual text-token states immediately before MMDiT joint attention, with coefficients that vary by block and denoising step.

\noindent \textbf{Understanding cross-modal interactions in diffusion models.}
Some works~\cite{hertz2022prompt,cao2023masactrl,liu2024towards} investigate the role of attention mechanisms in Unet-based diffusion models for instruction-based editing.
They reveal that attention values largely reflect the intensity of text-image interactions.
Classifier-free guidance (CFG)~\cite{cfg_ho2022classifier} adopts a static scaling factor to balance conditional and unconditional model outputs, thereby controlling textual strength.
Despite these advances, a holistic understanding of DiT remains underexplored.
Prior work~\cite{stableflow_avrahami2025,shin2025exploring,freeflux_wei2025} focused on adjusting attention mechanisms, evaluating rotary position embeddings, or analyzing specific blocks.
TACA~\cite{taca_lv2025rethinking} identified cross-modal attention imbalance and introduced a time-aware weighting scheme.
Li et al.~\cite{li2026unraveling} dissected the block contributions and proposed block-wise conditioning strategies.
TACA prescribes attention enhancement over time, while our coefficients are learned from data and can either enhance or suppress individual textual signals. CFG and its scheduling variants~\cite{cfg_ho2022classifier,cfg_scheduler_wang2024} adjust the overall sampling guidance, while our method acts inside the denoiser and can be combined with CFG.

\vspace{-4pt}
\section{Discussion}
\vspace{-4pt}
\label{sec:conclusion}

\begin{figure*}[t]
    \centering
    \includegraphics[width=0.9\textwidth]{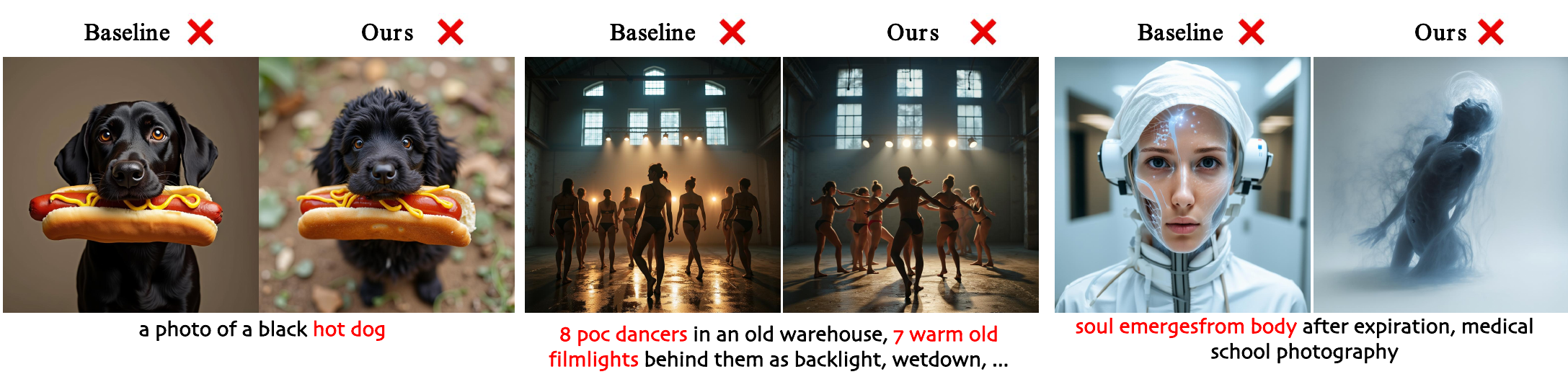}
    \vspace{-10pt}
    \caption{\textbf{Failure cases of \method.}
    Ambiguous concepts, precise counts and lighting constraints, and abstract descriptions remain challenging for both the baseline and \method.
    }
    \vspace{-10pt}
    \label{fig:main_failurecase}
\end{figure*}

\noindent \textbf{Limitation and future work.}
Since \method modulates textual interactions with a frozen backbone, it cannot reliably resolve ambiguous concepts, supply missing visual grounding, or enforce multiple simultaneous constraints shown in Fig.~\ref{fig:main_failurecase}.
Improved average alignment therefore does not guarantee prompt fidelity.
Future work will explore joint training with the backbone to improve visual grounding, explicit above constraints, and extensions to other attention architectures and modalities.

\noindent \textbf{Conclusions}. We presented \method, a plug-in module for rectified text-to-image diffusion transformers that learns to adaptively adjust block-wise interactions between token-level textual conditions across denoising trajectories.
Beyond improving the generation quality, \method provides insights into the dynamic interplay of cross-modal signals in transformer blocks and can be seamlessly applied to various tasks.
Our work benefits creative industries and accessibility tools, while it may also be misused to create realistic fake images, increasing risks of misinformation and deepfakes.

\newpage
\section*{Acknowledgements}

This work was supported in part by AI for Science Program, Shanghai Municipal Commission of Economy and Informatization (Grant No. 2025-GZL-RGZNBTBX-02017) and the National Natural Science Foundation of China (Grant No.62576103). The computations in this research were performed using the CFFF platform of Fudan University.

\bibliographystyle{plain}
\bibliography{references}

\newpage
\appendix

\section{Outline of Appendix}
\label{appendix:outline}

We provide additional details and results to supplement the main paper as follows.
We first present experimental setup and human-evaluation details in Sec.~\ref{appendix:experimental_setup}.
Sec.~\ref{appendix:sampling_trajectory} studies sampling trajectories and CFG robustness.
Sec.~\ref{appendix:sparse_enhance_results} compares static, enhancement-only, and full signed modulation.
Sec.~\ref{appendix:more_quantitative_results} provides additional baseline, alignment, efficiency, application, and ablation results.
Finally, Sec.~\ref{appendix:more_qualitative_results} provides token-level visualizations, further examples, and failure cases.

\section{Details of Experimental Setup}
\label{appendix:experimental_setup}

\subsection{Details of Baselines}
We conduct experiments on two strong text-to-image diffusion models: Stable Diffusion 3.5 Large (SD 3.5)~\cite{SD35_2024} and FLUX.1-Dev~\cite{flux_2024}.
We use the official checkpoints provided by the authors and do not perform any fine-tuning on these models.
During inference, model weights are loaded in 16-bit precision.
No acceleration techniques such as xformers or memory-efficient attention are used.
The Tab. \ref{tab:appendix_inference_parameter} summarizes the key hyperparameters used during inference for both models.

\begin{table}[!htbp]
    \small
    \centering
    \caption{Default inference hyper-parameters for both SD 3.5 and FLUX.}
    \label{tab:appendix_inference_parameter}
    \setlength{\tabcolsep}{2.5mm}{
    \begin{tabular}{lcccccc}
        \toprule
        Model & DiT Blocks & Parameters & Inference Timesteps & Sampler & CFG Scale & Image Size \\
        \midrule
        SD 3.5 & 38 & 8B & 28 & EDM & 7.0 & $1024 \times 1024$ \\
        FLUX & 57 & 12B & 28 & EDM & 3.5 & $1024 \times 1024$ \\
        \bottomrule
    \end{tabular}
    }
\end{table}

\subsection{Details of Evaluation Metrics}
To comprehensively evaluate the performance of our proposed method, we utilize a suite of established evaluation metrics that assess various aspects of text-to-image generation quality, including prompt adherence, image quality, and semantic alignment.
We employ the following evaluation metrics to assess the performance of our proposed method.

\begin{itemize}
    \item \textbf{HPSv2}~\cite{HPSV2_wu2023human} can effectively evaluate the preference score of text-image pairs or multiple images given a text prompt. It is trained on a large-scale human preference dataset and has shown strong correlation with human judgments.
    \item \textbf{HPSv3}~\cite{HPSV3_ma2025hpsv3} is derived from the QWen2-VL and trained on the HPDv3 dataset. It demonstrates improved performance over HPSv2 in evaluating text-image alignment and image quality.
    \item \textbf{ImageReward}~\cite{xu2023imagereward} utilizes preference data collected from human feedback and applies
    reinforcement learning to train a reward model that can evaluate the image preference score and provide feedback in reward signals.
    \item \textbf{Aesthetic Score}~\cite{laion_aesthetic_2022} primarily focuses on assessing the overall aesthetic quality of an image, which is built upon the CLIP models. This score is often subjective regarding composition, color harmony, and visual appearance.
    \item \textbf{Geneval Score}~\cite{geneval_ghosh2023} proposes an object-focused evaluation metric for text-to-image generation models. It defines a set of tasks to evaluate different aspects of text-image alignment, including ``Single Object'', ``Two Object'', ``Counting'', ``Position'', and ``Attribute binding''. The Geneval Score is computed as the average accuracy across these tasks. It primarily focused on evaluating the model's ability for semantic alignment.
    \item \textbf{CLIP Score}~\cite{CLIP_radford2021learning} measures the cosine similarity between the image and text embeddings obtained from a pre-trained CLIP model. CLIP-I and CLIP-T are used to evaluate the editing tasks.
    \item \textbf{T2I-CompBench}~\cite{t2i_compbench_huang2023t2i} measures compositional prompt following, including attributes and spatial relations.
    \item \textbf{DPG-Bench}~\cite{ella_hu2024} evaluates dense-prompt following across objects, attributes, and relationships.
\end{itemize}

\subsection{Details of Datasets}
\label{appendix:training_data}
\paragraph{Training Datasets.} HPDv3 dataset~\cite{HPSV3_ma2025hpsv3} totally contains $1.17$ million binary preference choices over $1.08$ million images, grouped in pairs by prompts.
These images are collected from various sources, including real images, MidJourney, Cogview4, FLUX, HunyuanDiT, Stable Diffusion 3 Medium, Stable Diffusion 1.4, and Glide, \textit{etc. }
Most sources are not competitive text-to-image models with FLUX and SD 3.5, which use all the images from suboptimal models, which may lead to inferior performance.

To ensure the training data quality, we filter the HPDv3 dataset using the following criteria: \textit{\textbf{1)}} Preference confidence filter. We only keep the image pairs with preference confidence scores higher than $0.9$. \textit{\textbf{2)}} Image quality filter. We only keep the image pairs where the preferred images are from real images, FLUX, or Kolors. After filtering, we obtain $77,252$ image pairs for training our \module on both baselines.

\paragraph{Subset of HPDv3 Benchmark.} HPDv3 benchmark consists of $12,000$ prompts covering a wide range of categories, including ``Animals'', ``Architecture'', ``Arts'', ``Characters'',``Design'', ``Food'', ``Nature Scenery'', ``Plants'', ``Products'', ``Science'', ``Transportation'', and ``Others''.
Each category contains $1,000$ prompts.
Tab. \ref{tab:hpsv3_results}, \ref{tab:more_metric_results} in the main paper report the results on the whole HPDv3 benchmark.
Other than the full benchmark, we also report the results on the $20\%$ subset of HPDv3 in Tab. \ref{tab:ablation_study} and \ref{tab:appendix_ablation_studies}, which contains $2,400$ prompts sampled from all categories.
The benchmark evaluation prompts are strictly excluded from the training subset.

\subsection{Effect of Training Data Quality}
\label{appendix:data_quality}
To isolate the role of filtering, we also train an Aid module using a mixture that includes lower-quality SD~1.5-generated images. Tab.~\ref{tab:data_quality} uses the same 2,400-prompt evaluation subset as the optimization ablations; the filtered-data SFT and DPO rows follow Tab.~\ref{tab:ablation_study}. Filtering improves the SFT HPSv3 result from $10.74$ to $10.84$, while the less-filtered Aid module still exceeds LoRA trained with filtered data. The comparison indicates that both data selection and the modulation design matter.

\begin{table}[htbp]
    \small
    \centering
    \caption{Training-data quality ablation on the 2,400-prompt HPDv3 subset. ``Filtered'' uses confidence $>0.9$ and preferred images from real images, FLUX, or Kolors; ``mixed'' additionally includes lower-quality SD~1.5 images.}
    \label{tab:data_quality}
    \setlength{\tabcolsep}{2.5mm}{
    \begin{tabular}{llcccc}
        \toprule
        Method & Training data & HPSv2 & HPSv3 & ImgRwd & Aes. \\
        \midrule
        FLUX & -- & 28.53 & 10.42 & 0.89 & 6.66 \\
        LoRA + SFT & Filtered & 28.54 & 10.43 & 0.86 & 6.63 \\
        \method + SFT & Mixed & 28.74 & 10.74 & 0.93 & 6.73 \\
        \method + SFT & Filtered & 28.80 & 10.84 & 0.95 & 6.76 \\
        \method + DPO & Filtered & 28.81 & 10.85 & 0.94 & 6.75 \\
        \bottomrule
    \end{tabular}
    }
\end{table}

\begin{figure}[t]
    \centering
    \includegraphics[width=0.95\linewidth]{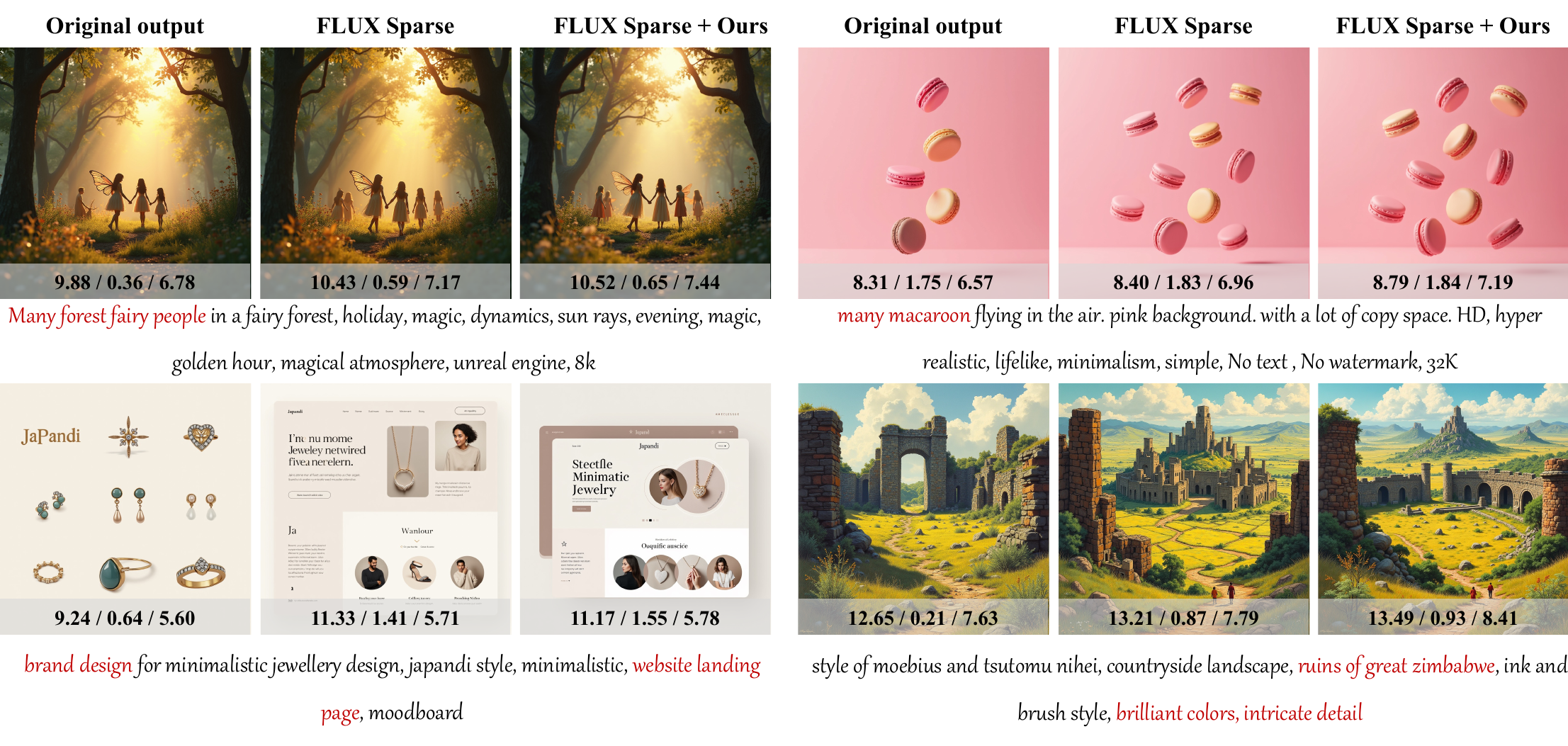}
    \caption{\textbf{Static sparse enhancement compared with full adaptive modulation and baseline FLUX.} The fixed block rule improves some prompts but cannot reproduce token- and timestep-dependent behavior. HPSv3, ImageReward, and Aesthetic scores are shown below each image.}
    \label{fig:appendix_sparse_enhance}
\end{figure}

\subsection{Details of optimization choices}
\label{appendix:implementation_details_of_optimization}

In Sec.~\ref{sec:our_designs} of the main paper, we introduce several optimization choices to enhance the training stability and performance of our \module.
In ablation studies (Tab.~\ref{tab:ablation_study}), we systematically evaluate the impact of four different optimization choices.
Here, we provide more implementation details for each choice.

\noindent \textbf{1) LoRA + SFT.}
We apply LoRA~\cite{lora_hu2022} with a rank of $64$ to all the linear layers in the FLUX. No \module is applied to the transformer blocks in this setting.

\noindent \textbf{2) LoRA + DPO.}
Based on the LoRA setting above, we further apply DPO to optimize the parameters to obtain better performance.
We set the DPO temperature $\beta$ to $0.1$ and use the original FLUX as the reference model.
The $\lambda_{dpo}$ is set to $1.0$, which is commonly used in previous works~\cite{dpo_wallace2024diffusion}.

\noindent \textbf{3) Ours + Rwd.}
In this setting, we use all the designs in our \module, including Tanh activation, $\alpha$ regularization ($\lambda_{reg}=0.1$), gated mechanism, and skipping during training ($p=0.1$).
The \module has an ``mlp\_hidden\_dim'' of $2048$.
We further incorporate the HPSv3 as the reward model to provide additional reward signals during training.
The reward weight $\lambda_{rwd}$ is set to $1.0$.
To keep the reward scores in a reasonable range, we only apply reward learning when the timestep $t$ is smaller than $0.3$.

\noindent \textbf{4) Ours + SFT.}
In this setting, we use all the designs in our \module, including Tanh activation, $\alpha$ regularization ($\lambda_{reg}=0.1$), gated mechanism, and skipping during training ($p=0.1$).
The \module has an ``mlp\_hidden\_dim'' of $2048$.
We apply the straightforward supervised fine-tuning (SFT) to optimize the parameters without additional reward signals or DPO.

\noindent \textbf{5) Ours + DPO.}
Similar to the settings of ``Ours + SFT'', we use all the designs in our \module.
We further apply DPO to optimize the parameters.
The DPO settings are the same as those in ``LoRA + DPO''.

In Tab.~\ref{tab:ablation_study}, we can observe that the giant leap in performance mainly comes from our \module designs and sparsity strategies, rather than the choice of optimization strategies.
Specifically, comparing ``LoRA + SFT'' with ``Ours + SFT'', our \module brings substantial improvements of $+0.41$ on HPSv3 and $+0.27$ on HPSv2, demonstrating the effectiveness of our adaptive modulation mechanism.
Incorporating additional optimization strategies like DPO or reward-based learning can achieve competitive performance and may provide marginal improvements on certain metrics.
However, the core contribution comes from our \module block designs, including the three-dimensional adaptive modulation, boundedness and sparsity constraints.
For optimization simplicity and to highlight the $\alpha$ learning mechanism, we adopt the straightforward SFT as our default optimization strategy.
Therefore, our default settings are using all the designs in our \module block with SFT optimization.

\section{Generalization to Different Sampling Trajectories}
\label{appendix:sampling_trajectory}

To validate that our learned \ module generalizes across different sampling trajectories and is not overfit to the specific 28-step EDM sampler used during training, we conduct two comprehensive experiments.

\subsection{Results on FLUX.1-Schnell}

We evaluate our \module modules on FLUX.1-Schnell, a distilled variant of FLUX.1 that uses only 4 sampling steps.
Despite the significant reduction in steps (from 28 to 4) and the different denoising scheme, our method still achieves substantial improvements over the baseline model.
As shown in Tab.~\ref{tab:appendix_schnell_results}, our \module modules bring consistent gains across all metrics.
This demonstrates that the learned $\alpha$ values are not overfit to the specific 28-step trajectory and can generalize to drastically different sampling schemes.

\begin{table}[htbp]
    \small
    \centering
    \caption{\textbf{Quantitative results on FLUX.1-Schnell with 4 sampling steps.}
    Our method achieves consistent improvements even with a distilled model using only 4 steps, demonstrating strong generalization across different sampling trajectories.
    }
    \label{tab:appendix_schnell_results}
    \setlength{\tabcolsep}{3.5mm}{
    \begin{tabular}{lcccc}
        \toprule
        Models & HPSv2 & HPSv3 & ImgRwd & Aes. \\
        \midrule
        FLUX.1-Schnell & 28.48 & 9.95 & 0.87 & 6.55 \\
        +Ours & \textbf{28.65} & \textbf{10.14} & \textbf{0.91} & \textbf{6.64} \\
        \bottomrule
    \end{tabular}
    }
\end{table}

\subsection{Comprehensive Evaluation Across Different Timesteps}

We conduct a systematic evaluation on FLUX.1-Dev across 6 different sampling steps: $\{4, 8, 10, 15, 28, 50\}$ using $1,000$ prompts with fixed random seeds.
The results are summarized in Tab.~\ref{tab:appendix_different_timesteps}.
Our method consistently outperforms the baseline across all timestep configurations.
Notably, at 4 steps, our method achieves $+1.14$ HPSv2, $+3.47$ HPSv3, and $+0.56$ ImageReward improvements, demonstrating robust generalization despite the dramatic trajectory shift.
At $50$ steps, the improvements remain consistent.
These results confirm that our \module modules do not overfit to the specific denoising trajectory and maintains effectiveness across a wide range of sampling configurations.

\begin{table}[htbp]
    \small
    \centering
    \caption{\textbf{Comprehensive evaluation across different sampling steps on FLUX.1-Dev.}
    Our method consistently outperforms the baseline across all timestep configurations, demonstrating robust generalization to different sampling trajectories.
    All experiments are conducted on $1,000$ prompts with fixed random seeds.
    }
    \label{tab:appendix_different_timesteps}
    \setlength{\tabcolsep}{1.5mm}{
    \begin{tabular}{llcccccc}
        \toprule
        Metric & Method & 4 & 8 & 10 & 15 & 28 & 50 \\
        \midrule
        \multirow{2}{*}{HPSv2} & Baseline & 25.48 & 27.53 & 27.82 & 28.44 & 28.95 & 29.11 \\
         & Ours & \textbf{26.62} & \textbf{28.29} & \textbf{28.55} & \textbf{28.95} & \textbf{29.15} & \textbf{29.23} \\
        \midrule
        \multirow{2}{*}{HPSv3} & Baseline & 5.07 & 9.70 & 9.94 & 10.45 & 10.89 & 11.07 \\
         & Ours & \textbf{8.54} & \textbf{11.31} & \textbf{11.29} & \textbf{11.34} & \textbf{11.33} & \textbf{11.41} \\
        \midrule
        \multirow{2}{*}{ImgRwd} & Baseline & 0.21 & 0.87 & 0.90 & 1.01 & 1.08 & 1.10 \\
         & Ours & \textbf{0.77} & \textbf{1.10} & \textbf{1.12} & \textbf{1.14} & \textbf{1.14} & \textbf{1.13} \\
        \midrule
        \multirow{2}{*}{Aes.} & Baseline & 5.95 & 6.67 & 6.70 & 6.84 & 6.88 & 6.88 \\
         & Ours & \textbf{6.62} & \textbf{7.03} & \textbf{7.02} & \textbf{7.02} & \textbf{6.96} & \textbf{6.93} \\
        \bottomrule
    \end{tabular}
    }
\end{table}

These experiments provide strong evidence that our training does not modify the noise schedule or denoising scheme, and the learned $\alpha$ values are theoretically applicable to all timesteps.
This is further confirmed by Figure~\ref{fig:alpha_distribution_different_timesteps} in Sec.~\ref{appendix:alpha_distribution_different_timesteps}, where we visualize the $\alpha$ mean values across different sampling timesteps.
The curves converge to a stable pattern, demonstrating that \module does not overfit to the specific 28-step trajectory.

\subsection{Interaction with CFG Scale}
\label{appendix:cfg_sweep}
CFG rescales the sampling prediction globally, whereas \method changes text-token states before joint attention. We evaluate both methods with the same prompts and seeds at seven FLUX guidance scales. Tab.~\ref{tab:cfg_sweep} shows a GenEval improvement at every tested scale; the mean gain is $0.038$. The result supports complementarity within the tested range, rather than invariance to arbitrary guidance choices.

\begin{table}[htbp]
    \small
    \centering
    \caption{GenEval across FLUX CFG scales with fixed prompts and seeds.}
    \label{tab:cfg_sweep}
    \setlength{\tabcolsep}{2mm}{
    \begin{tabular}{lccccccc}
        \toprule
        CFG scale & 1 & 2 & 4 & 6 & 8 & 10 & 15 \\
        \midrule
        FLUX & 0.618 & 0.653 & 0.662 & 0.667 & 0.664 & 0.638 & 0.586 \\
        FLUX + \method & \textbf{0.652} & \textbf{0.674} & \textbf{0.690} & \textbf{0.703} & \textbf{0.705} & \textbf{0.682} & \textbf{0.647} \\
        \bottomrule
    \end{tabular}
    }
\end{table}

\section{Sparse Enhancement Results}
\label{appendix:sparse_enhance_results}

Li et al.~\cite{li2026unraveling} selected a subset of transformer blocks in MMDiT to enhance the text embeddings.
This approach depends on the prior knowledge of which blocks are vital for text-image generation.
From Fig.~\ref{fig:alpha}(a) in the main paper, we can observe that some blocks have higher $\alpha$ values, indicating their importance in text-image interactions.
We simply select the blocks $\{1,15,36,41,48\}$ for FLUX and set the $\alpha$ to $0.5$ for these blocks and $0.0$ for others.
This sparse enhancement strategy overlooks the learned timestep-awareness and token-level modulation in our \module.
However, it provides insights into the effectiveness of enhancing only a few crucial blocks.
As shown in Fig.~\ref{fig:appendix_sparse_enhance} and Tab.~\ref{tab:appendix_sparse_enhance_results}, static enhancement can recover some important content but falls below full adaptive modulation on HPSv3.
ImageReward score ($0.98$) of sparse enhancement even outperforms our full method.
This shows that the selected blocks are useful, while their fixed amplification does not account for token, timestep, or sign differences.
The comparison supports the functional relevance of the learned block pattern without attributing the full method's gains to block selection alone.

\begin{table}[htbp]
    \small
    \centering
    \caption{\textbf{Quantitative results of sparse enhancement strategy compared to the full method and baseline FLUX.}
    Sparse enhancement improves on the baseline but remains below full modulation on HPSv3.
    This indicates the effectiveness of our learned $\alpha$ values in identifying important blocks for text-image interactions.
    }
    \label{tab:appendix_sparse_enhance_results}
    \setlength{\tabcolsep}{3.5mm}{
    \begin{tabular}{lcccc}
        \toprule
        Strategies & HPSv2 & HPSv3 & ImgRwd & Aes. \\
        \midrule
        FLUX & 28.53 & 10.42 & 0.89 & 6.66 \\
        Sparse Enhancement & \underline{28.61} & \underline{10.57} & \textbf{0.98} & \underline{6.71} \\
        Our Full Method& \textbf{28.80}  & \textbf{10.84} & \underline{0.95} & \textbf{6.76} \\
        \bottomrule
    \end{tabular}
    }
\end{table}

\noindent \textbf{Signed modulation versus enhancement alone.}
The fixed sparse rule above only increases selected block signals. In a separate lightweight ($d=512$) comparison, we also clamp the learned coefficient to $\max(\alpha,0)$ at inference. Tab.~\ref{tab:signed_modulation_control} shows that adaptive enhancement improves on fixed scaling, while restoring negative coefficients improves both metrics further. This supports a functional role for suppression beyond the visible negative regions in Fig.~\ref{fig:alpha}.

\begin{table}[htbp]
    \small
    \centering
    \caption{Static, enhancement-only, and signed adaptive controls in the additional lightweight ($d=512$) FLUX comparison. The default $d=2048$ result is reported separately in Tab.~\ref{tab:appendix_sparse_enhance_results}.}
    \label{tab:signed_modulation_control}
    \setlength{\tabcolsep}{3mm}{
    \begin{tabular}{lcc}
        \toprule
        Control & HPSv3 & GenEval \\
        \midrule
        FLUX baseline & 10.42 & 0.676 \\
        Sparse static scaling & 10.57 & 0.683 \\
        Enhancement-only, $\alpha\leftarrow\max(\alpha,0)$ & 10.72 & 0.692 \\
        Full signed \method ($d=512$) & \textbf{10.82} & \textbf{0.709} \\
        \bottomrule
    \end{tabular}
    }
\end{table}

\section{More Quantitative Results}
\label{appendix:more_quantitative_results}

\subsection{Compositional and Text-Image Alignment Benchmarks}
\label{appendix:alignment_benchmarks}
The preference-oriented scores in the main paper are complemented by compositional and dense-prompt evaluations. For each backbone, baseline and \method generations use matching prompts and seeds. Tab.~\ref{tab:alignment_benchmarks} shows gains on both backbones for T2I-CompBench~\cite{t2i_compbench_huang2023t2i}, DPG-Bench~\cite{ella_hu2024}, and CLIP Score~\cite{CLIP_radford2021learning}. The small CLIP Score differences are reported without claiming that this metric alone establishes semantic correctness.

\begin{table}[htbp]
    \small
    \centering
    \caption{Additional alignment results. Higher is better for all three metrics.}
    \label{tab:alignment_benchmarks}
    \setlength{\tabcolsep}{3mm}{
    \begin{tabular}{lccc}
        \toprule
        Model & T2I-CompBench & DPG-Bench & CLIP Score \\
        \midrule
        FLUX & 0.4739 & 83.67 & 28.13 \\
        FLUX + \method & \textbf{0.5006} & \textbf{87.14} & \textbf{28.35} \\
        SD~3.5 & 0.4980 & 76.40 & 27.88 \\
        SD~3.5 + \method & \textbf{0.5249} & \textbf{79.32} & \textbf{27.95} \\
        \bottomrule
    \end{tabular}
    }
\end{table}

\subsection{Additional Guidance, PEFT, and Reward-Guided Baselines}
\label{appendix:additional_baselines}
We compare further FLUX adaptations of sampling guidance~\cite{sag_hong2023,pag_ahn2024,cfg_scheduler_wang2024}, parameter-efficient tuning~\cite{lora_hu2022,dora_liu2024}, and reward-guided tuning~\cite{draft_clark2024,alignprop_prabhudesai2023}. Tab.~\ref{tab:additional_baselines} uses the 2,400-prompt HPSv3 subset and the lightweight $d=512$ \method variant ($+0.38\%$ parameters), rather than the default $d=2048$ variant ($+1.51\%$) in the main paper. The reward-guided baselines approach its HPSv3 score, whereas the lightweight \method gives the highest GenEval in this comparison. Tab.~\ref{tab:hpsv3_results} separately reports the default model on the full 12,000-prompt benchmark.

\begin{table}[htbp]
    \small
    \centering
    \caption{Additional FLUX comparison on the 2,400-prompt HPSv3 subset and GenEval, using the lightweight \method variant. Guidance methods add no trainable parameters; parameter percentages are relative to FLUX.}
    \label{tab:additional_baselines}
    \setlength{\tabcolsep}{3mm}{
    \begin{tabular}{llccc}
        \toprule
        Method & Type & Extra params & HPSv3 & GenEval \\
        \midrule
        FLUX & Baseline & -- & 10.42 & 0.676 \\
        SAG~\cite{sag_hong2023} & Guidance & 0 & 10.45 & 0.678 \\
        PAG~\cite{pag_ahn2024} & Guidance & 0 & 10.51 & 0.678 \\
        Dynamic CFG~\cite{cfg_scheduler_wang2024} & Guidance & 0 & 10.50 & 0.681 \\
        LoRA~\cite{lora_hu2022} & PEFT & $+0.31\%$ & 10.43 & 0.683 \\
        DoRA~\cite{dora_liu2024} & PEFT & $+0.33\%$ & 10.47 & 0.691 \\
        DRaFT~\cite{draft_clark2024} & Reward-guided & $+0.31\%$ & 10.81 & 0.664 \\
        AlignProp~\cite{alignprop_prabhudesai2023} & Reward-guided & $+0.31\%$ & \textbf{10.82} & 0.670 \\
        \method ($d=512$) & Adaptive modulation & $+0.38\%$ & \textbf{10.82} & \textbf{0.709} \\
        \bottomrule
    \end{tabular}
    }
\end{table}

\subsection{Parameter Efficiency and Computational Overhead Analysis}
\label{appendix:parameter_efficiency}

To comprehensively evaluate the efficiency of our proposed \method, we conduct a detailed comparison with LoRA-based methods under similar parameter budgets.
We measure the exact computational overhead on $1,000$ images of resolution $1024 \times 1024$ with 28 sampling steps on FLUX.1-Dev.
The evaluation includes additional parameters, peak VRAM usage, average inference time per image, and performance metrics across HPSv2, HPSv3, ImageReward, and Aesthetic Score.
We compare our method with two LoRA configurations: LoRA with rank $32$ and rank $128$, as well as two variants of our Aid module with hidden dimensions of $512$ and $2048$.
The results are summarized in Tab.~\ref{tab:appendix_parameter_efficiency}.

\begin{table}[htbp]
    \small
    \centering
    \caption{\textbf{Comparison of parameter efficiency and computational overhead between LoRA and our method.}
    Runtime is measured on $1,000$ FLUX.1-Dev images at $1024 \times 1024$ resolution and 28 sampling steps; preference scores use the 2,400-prompt HPDv3 subset.
    Our method achieves substantially better performance than LoRA at comparable parameter budgets.
    }
    \label{tab:appendix_parameter_efficiency}
    \setlength{\tabcolsep}{1.8mm}{
    \begin{tabular}{lccccccc}
        \toprule
        Method & \makecell{Additional\\Param (M)} & \makecell{Peak\\VRAM (GB)} & \makecell{Avg.\\Time (s)} & HPSv2 & HPSv3 & ImgRwd & Aes. \\
        \midrule
        FLUX.1-Dev & — & 37.22 & 5.7483 & 28.53 & 10.42 & 0.89 & 6.66 \\
        \midrule
        +LoRA (rank=32) & 37.35 (+0.31\%) & 37.67 & 5.7649 & 28.54 & 10.43 & 0.86 & 6.63 \\
        +LoRA (rank=128) & 149.42 (+1.26\%) & 37.96 & 5.8164 & 28.55 & 10.44 & 0.88 & 6.68 \\
        \midrule
        +Ours (dim=512) & 45.21 (+0.38\%) & 37.83 & 5.7897 & 28.80 & 10.82 & 0.95 & 6.75 \\
        +Ours (dim=2048) & 179.77 (+1.51\%) & 38.08 & 5.8252 & 28.80 & 10.84 & 0.95 & 6.76 \\
        \bottomrule
    \end{tabular}
    }
\end{table}

\noindent \textbf{Parameter Efficiency.}
At comparable parameter budgets, our Aid module demonstrates substantially superior performance over LoRA-based methods.
Specifically, when comparing LoRA (rank=128) with 149.42M additional parameters (+1.26\%) against our method (dim=2048) with 179.77M parameters (+1.51\%), our approach achieves significantly larger improvements: +0.42 on HPSv3 and +0.27 on HPSv2, while LoRA only gains +0.02 on both metrics.
Even our lightweight variant (dim=512) with only 45.21M parameters (+0.38\%) substantially outperforms both LoRA configurations, achieving +0.40 on HPSv3 and +0.27 on HPSv2.
This demonstrates that the performance gains stem from our Aid module's token-adaptive modulation design rather than simply increasing parameter count.

\noindent \textbf{Computational Overhead.}
Our method introduces modest computational overhead that is comparable to LoRA at similar parameter scales.
The default configuration (dim=2048) adds only 0.86 GB of peak VRAM and incurs a wall-clock overhead of just +1.34\% (0.077s per image) compared to the baseline.
This overhead is comparable to LoRA (rank=128), which adds 0.74 GB VRAM and +1.18\% inference time, yet our method delivers substantially larger performance gains.
The lightweight variant (dim=512) further reduces the overhead to +0.61 GB VRAM and +0.72\% inference time with minimal performance degradation, demonstrating the flexibility of our approach to balance between efficiency and performance.

\noindent \textbf{Architectural Advantages over LoRA.}
The superior performance of our Aid module over LoRA at comparable parameter budgets highlights the importance of where and how modulation is applied.
Unlike LoRA, which uniformly adjusts backbone weights across all layers, our Aid module provides dynamic, input-dependent modulation of text features before the joint self-attention in MMDiT, enabling fine-grained control over text-image interactions at the token level.
This architectural design allows our method to adaptively modulate text conditioning strength across blocks, timesteps, and tokens, addressing the token imbalance problem that LoRA cannot capture.
The consistent improvements across all metrics (HPSv2, HPSv3, ImageReward, and Aesthetic Score) further validate that our Aid module's structured modulation capability is more effective than LoRA's uniform weight adaptation for enhancing text-image alignment.

\subsection{Quantitative Results on Applications}
\label{sec:appendix_applications}

In Sec.~\ref{sec:applications} of the main paper, we demonstrate the versatility of our proposed \method by applying it to different conditional generation tasks.
We visualize some qualitative results in Fig.~\ref{fig:editing_qualitative}, Fig.~\ref{fig:ctrl_qualitative} and Fig.~\ref{fig:lora_qualitative}.
Here, we further evaluate our method on two applications: 1) using ``Canny''\footnote{https://huggingface.co/black-forest-labs/FLUX.1-Canny-dev} and ``Depth''\footnote{https://huggingface.co/black-forest-labs/FLUX.1-Depth-dev} conditions. 2) using ``Sketch''\footnote{https://huggingface.co/Shakker-Labs/FLUX.1-dev-LoRA-Children-Simple-Sketch} and ``Glass''\footnote{https://civitai.com/models/553811/stained-glass?modelVersionId=771546} LoRA modules to adapt to new styles.
The quantitative results are shown in Tab.~\ref{tab:appendix_application_results}.

\begin{table}[htbp]
    \small
    \centering
    \caption{\textbf{Quantitative results of applying our \method to style LoRAs and control conditions generation.}
    Our method can be flexibly applied to different applications while obtaining competitive performance improvements.
    }
    \label{tab:appendix_application_results}
    \setlength{\tabcolsep}{2.5mm}{
    \begin{tabular}{lcccc}
        \toprule
        Strategies & HPSv2 & HPSv3 & ImgRwd & Aes. \\
        \midrule
        LoRA & 28.50 & 10.38 & 0.87 & 6.64 \\
        LoRA + Ours& 28.54  & 10.43 & 0.86 & 6.63 \\
        \midrule
        Control & 28.57 & 10.48 & 0.90 & 6.67 \\
        Control + Ours & 28.76 & 10.68 & 0.93 & 6.70 \\
        \bottomrule
    \end{tabular}
    }
\end{table}

These results are obtained on the FLUX model on the $20\%$ of the HPDv3 dataset.
Our method not only seamlessly integrates with these models but also significantly enhances their performance in terms of prompt adherence and image quality.
This demonstrates the versatility and effectiveness of our approach across different conditional generation tasks.

\subsection{Heatmap of Alpha Values}
\label{appendix:heatmap_alpha}

\begin{figure}[htbp]
    \centering
    \begin{subfigure}{0.49\linewidth}
        \centering
        \setlength{\abovecaptionskip}{0pt}
        \setlength{\belowcaptionskip}{3pt}
        \includegraphics[width=\linewidth]{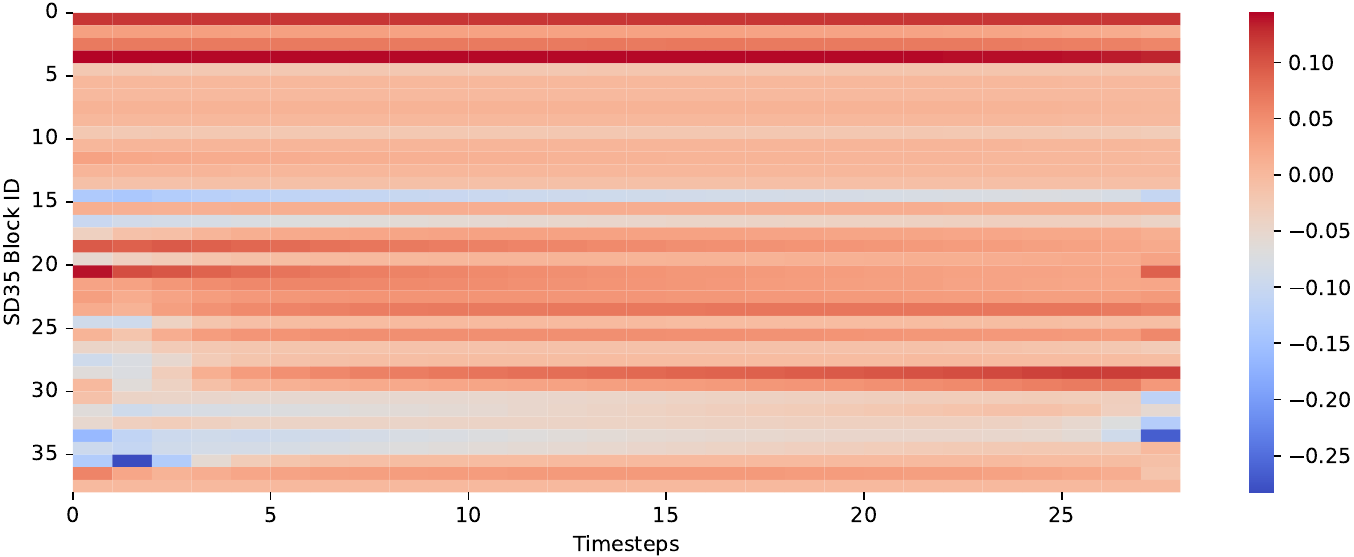}
        \caption{Block-timestep $\alpha$ heatmap of SD 3.5.}
        \label{fig:alpha_timestep_heatmap_sd35}
    \end{subfigure}
    \begin{subfigure}{0.49\linewidth}
        \centering
        \setlength{\abovecaptionskip}{0pt}
        \setlength{\belowcaptionskip}{3pt}
        \includegraphics[width=\linewidth]{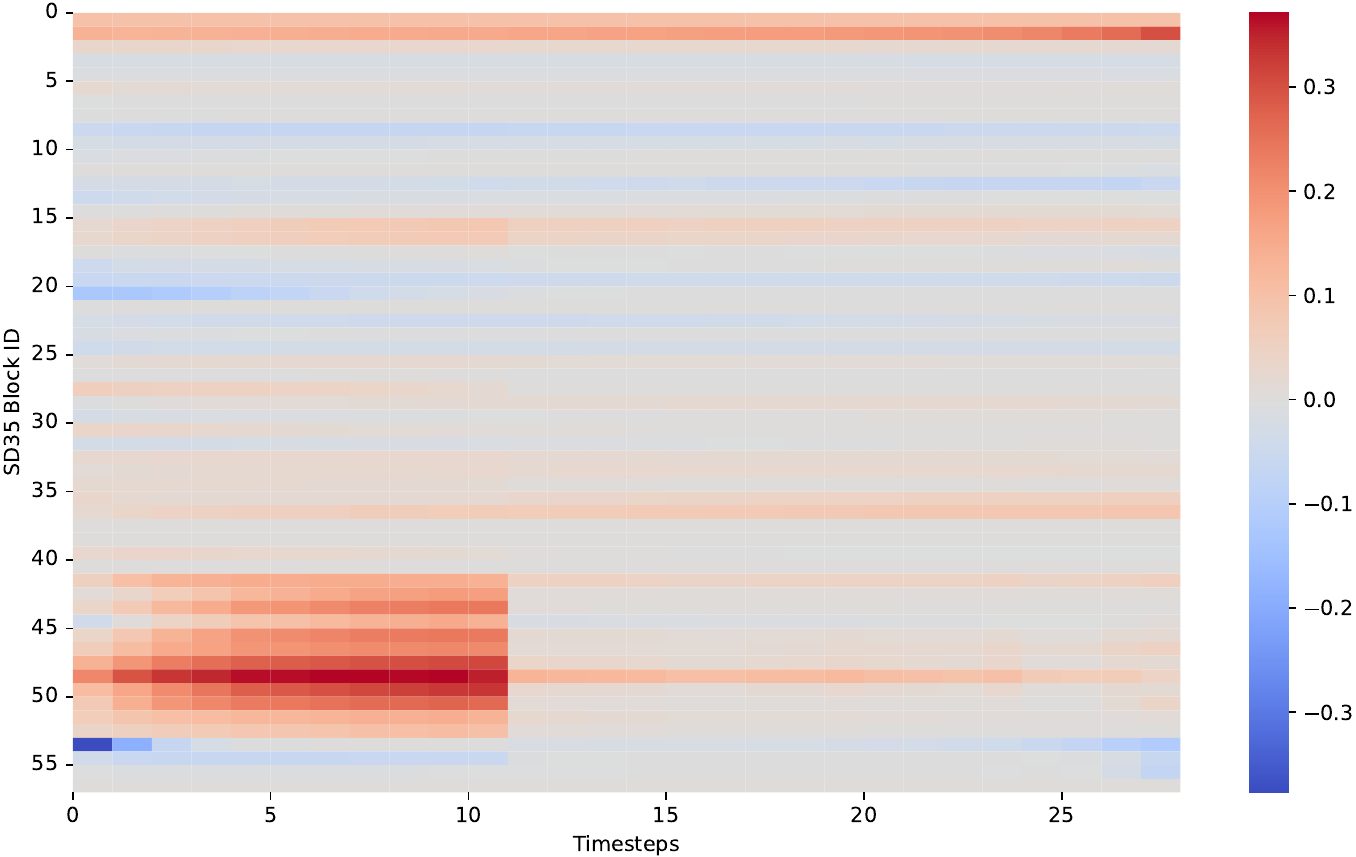}
        \caption{Block-timestep $\alpha$ heatmap of FLUX.}
        \label{fig:alpha_timestep_heatmap_flux}
    \end{subfigure}

    \caption{Visualization of the $\alpha$ across different timesteps and blocks for the models.}

    \label{fig:alpha_timestep_heatmap}
\end{figure}

Fig.~\ref{fig:alpha}(a) and (b) in the main paper show the distribution of $\alpha$ values across various transformer blocks and timesteps, respectively.
Here, Fig.~\ref{fig:alpha_timestep_heatmap} provides a more detailed visualization of each block's $\alpha$ values at different timesteps for both models.
Actually, the sharp changes described in Fig.~\ref{fig:alpha}(c) of the main paper can also be observed in the attention heatmap in Fig.~\ref{fig:alpha_timestep_heatmap_flux}, where some blocks ($41st-54th$) exhibit significant variations in $\alpha$ values at the timestep $11$.
In this way, we can clearly see how our \method dynamically adjusts the block-wise interactions between token-level textual conditions across different timesteps.

\subsection{Distribution of $\alpha$ across different timesteps}
\label{appendix:alpha_distribution_different_timesteps}
\begin{figure}[htbp]
    \centering
    \includegraphics[width=\linewidth]{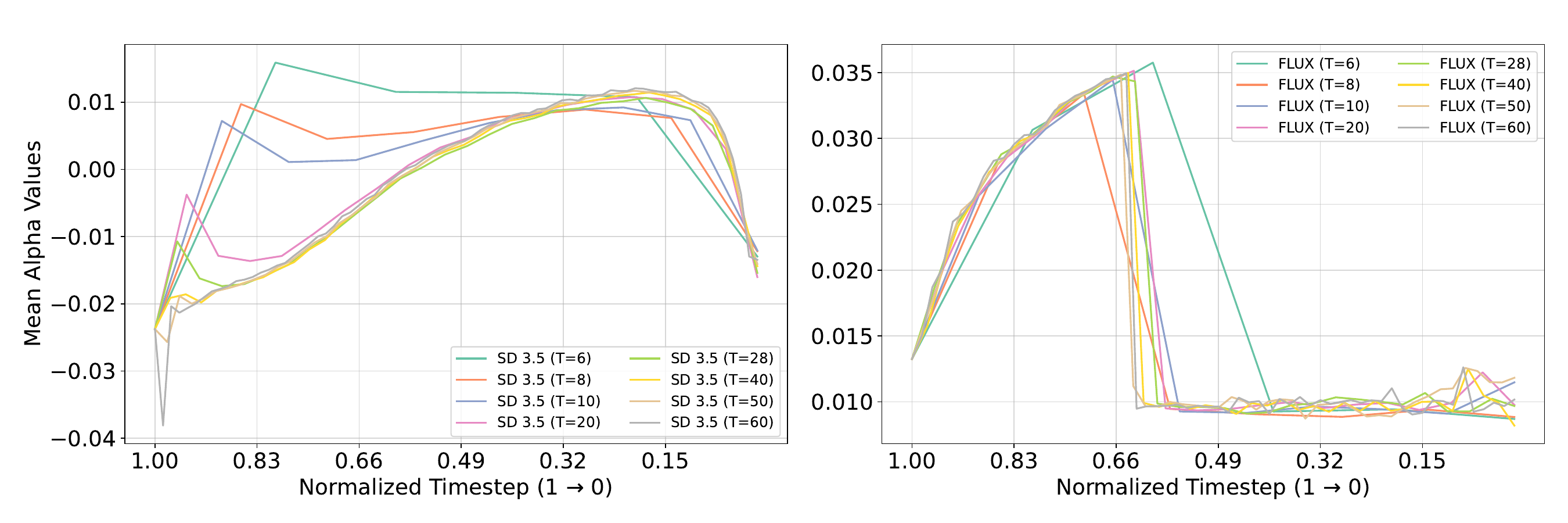}
    \caption{\textbf{Distribution of $\alpha$ across different timesteps for both SD 3.5 and FLUX.}
    The distributions of $\alpha$ values at different timesteps exhibit similar patterns and do not vary significantly with the number of sampling steps.
    }
    \label{fig:alpha_distribution_different_timesteps}
\end{figure}

In Fig.~\ref{fig:alpha}(c) of the main paper, we plot the distribution of $\alpha$ values across timesteps ($T=28$) for both SD 3.5 and FLUX.
Here, we further visualize the distribution of $\alpha$ values across different timesteps ($T \in \{6,8,10,20,28,40,50,60\}$) for both models in Fig.~\ref{fig:alpha_distribution_different_timesteps}.
As shown in Fig.~\ref{fig:alpha_distribution_different_timesteps}, the distributions of $\alpha$ values at different timesteps exhibit similar patterns and do not vary significantly with the number of sampling steps.
As the timestep increases, the curve shows a stable convergence trend, indicating that the modulation ability of our \method stabilizes at higher timesteps.
This indicates that our \module effectively captures the timestep-aware interactions between token-level textual conditions, maintaining consistent behavior across various sampling step configurations.

\subsection{Isolating the Contribution of Timestep, Block, and Token Dimensions}
\label{appendix:dimension_ablation}
To quantify the individual contribution of each dimension in our three-dimensional adaptive modulation design, we conduct controlled ablation studies by removing the adaptivity of one dimension while keeping the other two fully adaptive.
Specifically, we evaluate three variants:
\textbf{(1) Share Blocks}: A single \module is shared across all transformer blocks, removing block-wise adaptivity.
\textbf{(2) Share Tokens}: A scalar $\alpha$ is shared across all text tokens, removing token-level adaptivity.
\textbf{(3) Timestep variants}: We test three strategies to remove timestep adaptivity: \textit{mean}—averaging $\alpha$ across all timesteps; \textit{1st hack}—fixing $\alpha$ to values from the 1st denoising step; \textit{10th hack}—fixing $\alpha$ to values from the 10th denoising step.

As shown in Tab.~\ref{tab:dimension_ablation} in the main paper, sharing the module across blocks causes the largest HPSv3 drop ($-1.26$). Sharing one coefficient across prompt tokens reduces HPSv3 by $0.56$. Averaging $\alpha$ over timesteps reduces it by $0.38$, leaving only a $0.04$ gain over FLUX; fixing it to the first or tenth step lowers the score below FLUX. These controls show why the three adaptive dimensions are retained together.

\begin{table}[htbp]
    \small
    \centering
    \caption{\textbf{Contribution of three dimension adaptivity on FLUX.} }
    \label{tab:dimension_ablation}
    \setlength{\tabcolsep}{3.0mm}{
    \begin{tabular}{lcccc}
        \toprule
        Setting & HPSv2 & HPSv3 & ImgRwd & Aes. \\
        \midrule
        FLUX baseline & 28.53 & 10.42 & 0.89 & 6.66 \\
        \method (full) & \textbf{28.80} & \textbf{10.84} & \textbf{0.95} & \textbf{6.76} \\
        \midrule
        Shared across blocks & 27.41 & 9.58 & 0.77 & 6.34 \\
        Shared across tokens & 28.07 & 10.28 & 0.88 & 6.48 \\
        \midrule
        Time-averaged $\alpha$ & 28.63 & 10.46 & 0.91 & 6.53 \\
        Fixed 1st-step $\alpha$ & 28.56 & 10.35 & 0.82 & 6.33 \\
        Fixed 10th-step $\alpha$ & 28.53 & 10.32 & 0.82 & 6.35 \\
        \bottomrule
    \end{tabular}
    }
\end{table}

\subsection{Effect of $L_2$ Regularization on Magnitude and Stability}
\label{appendix:regularization_effect}

\begin{figure}[htbp]
	\centering
	\includegraphics[width=0.65\linewidth]{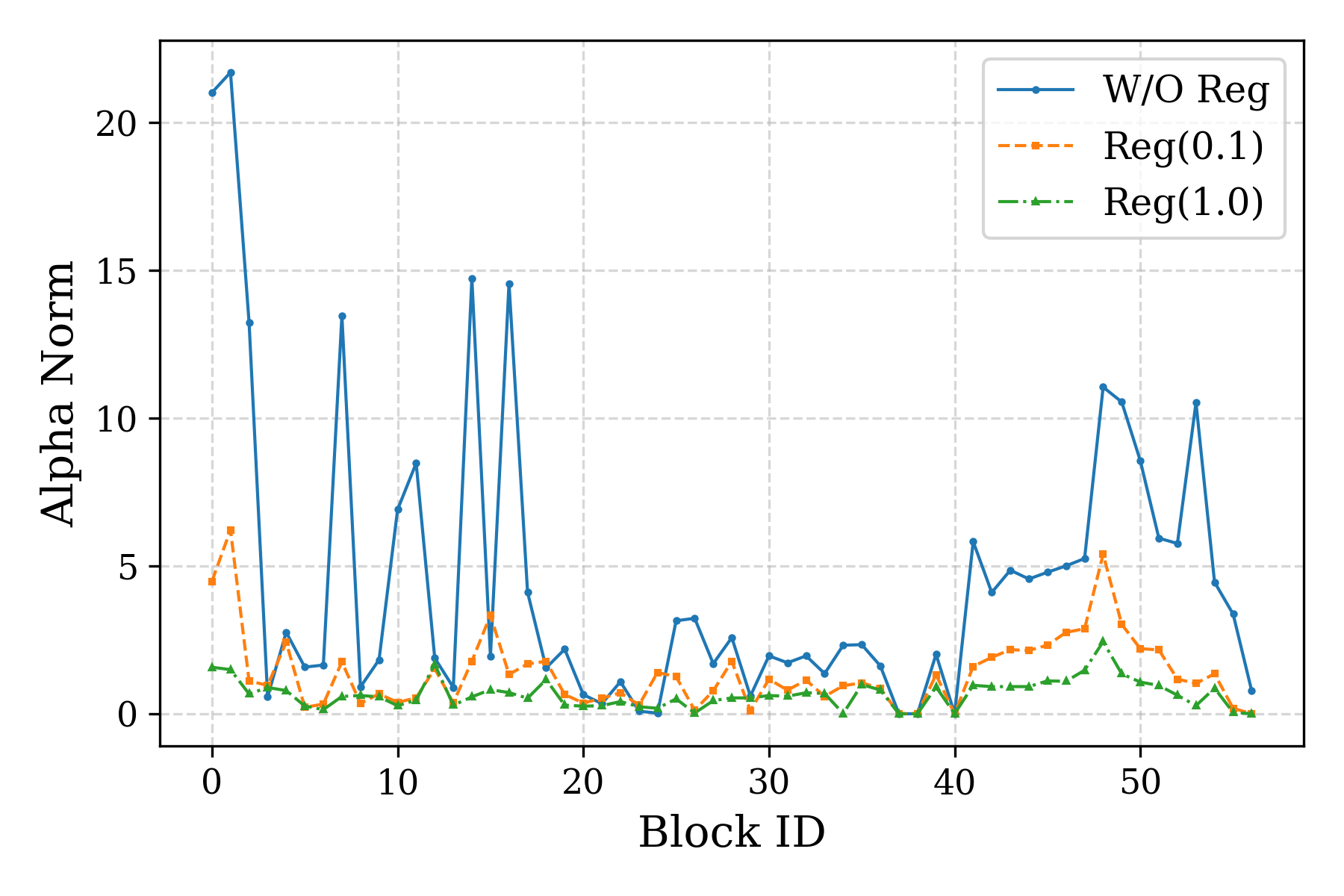}
	\caption{\textbf{Effect of regularization on $\alpha$ distribution across transformer blocks.}
	Without regularization (left), $\alpha$ values show erratic fluctuations and extreme values.
    With regularization (right), $\alpha$ magnitudes become smaller and more stable, maintaining interpretable modulation patterns.
	}
	\label{fig:supp_curve_regs}
\end{figure}

To demonstrate the importance of regularization in our \module design, we analyze how the $L_2$-norm term $\mathcal{L}_{reg}=\mathbb{E}_{t,l}[\|\alpha_t^l\|_2]$ affects the learned $\alpha$ values.
The role of $\mathcal{L}_{reg}$ is not merely to maximize performance metrics, but to keep $\alpha$ within a reasonable range and ensure training stability.
As shown in Fig.~\ref{fig:supp_curve_regs}, we compare the distribution of $\alpha$ values across transformer blocks with and without regularization.

Without regularization, $\alpha$ exhibits erratic fluctuations and extreme values across blocks, causing unstable modulation that undermines interpretable patterns.
In contrast, with regularization ($\lambda_{reg}=0.1$), the $\alpha$ magnitudes become smaller and more stable, maintaining meaningful token-level patterns while preventing excessive modulation that would distort text feature distributions. The gate separately encourages selective modulation.
This demonstrates that regularization is essential for achieving both effective and interpretable modulation in our \module.

\subsection{More Ablation of dropout probability.}
\label{appendix:dropout_prob}
We further investigate the impact of different dropout probabilities in our \module on the performance of FLUX.
As shown in Tab. \ref{tab:appendix_ablation_studies}, we experiment with dropout probabilities ranging from $0.0$ to $0.7$.
The results indicate that a dropout probability of $0.1$ yields the best performance across all metrics.
When the dropout probability grows to $0.2$, the performance slightly decreases but remains competitive.
Big probabilities like $0.5$ and $0.7$ lead to significant drops in all metrics.
These results suggest that incorporating a small amount of dropout helps regularize the model, preventing overfitting while still allowing it to learn effective modulation patterns.
In contrast, big $p$ values may cause excessive information loss during training and inference mismatch, leading to degraded performance.

\begin{table}[!htbp]
    \small
    \centering
    \caption{More ablation studies on different dropout probabilities.}
    \label{tab:appendix_ablation_studies}
    \setlength{\tabcolsep}{2.5mm}{
    \begin{tabular}{l|lcccc}
        \toprule
        Categories & Settings & HPSv2 & HPSv3 & ImgRwd & Aes. \\
        \midrule
        \multicolumn{2}{c}{FLUX} & 28.53 & 10.42 & 0.89 & 6.66 \\
        \midrule
        \multirow{6}{*}{\begin{tabular}[c]{@{}c@{}} Dropout \\ Probability \end{tabular}} & $p$=0.0 & 28.71 & 10.74 & \underline{0.94} & 6.72 \\
          & $p$=0.1 & \textbf{28.80} & \textbf{10.84} & \textbf{0.95} & \textbf{6.76} \\
          & $p$=0.2 & \underline{28.73} & \underline{10.77} & \underline{0.94} & \underline{6.73} \\
          & $p$=0.3 & 28.47 & 10.44 & 0.87 & 6.64 \\
          & $p$=0.5 & 28.21 & 10.19 & 0.77 & 6.59 \\
          & $p$=0.7 & 27.03 & 9.42 & 0.35 & 6.54 \\
        \bottomrule
    \end{tabular}
    }
\end{table}

\section{More Qualitative Results}
\label{appendix:more_qualitative_results}

\subsection{Visualization of Adaptive Modulation Patterns: A Case Study}
\label{appendix:adaptive_modulation_visualization}

To provide fine-grained interpretability of our Aid module, we present a concrete case study tracing per-token $\alpha$ dynamics for a complex prompt: \textit{``a red cat sitting on a blue chair in a sunny garden''}.

\begin{figure*}[h]
    \centering
    \includegraphics[width=0.95\linewidth]{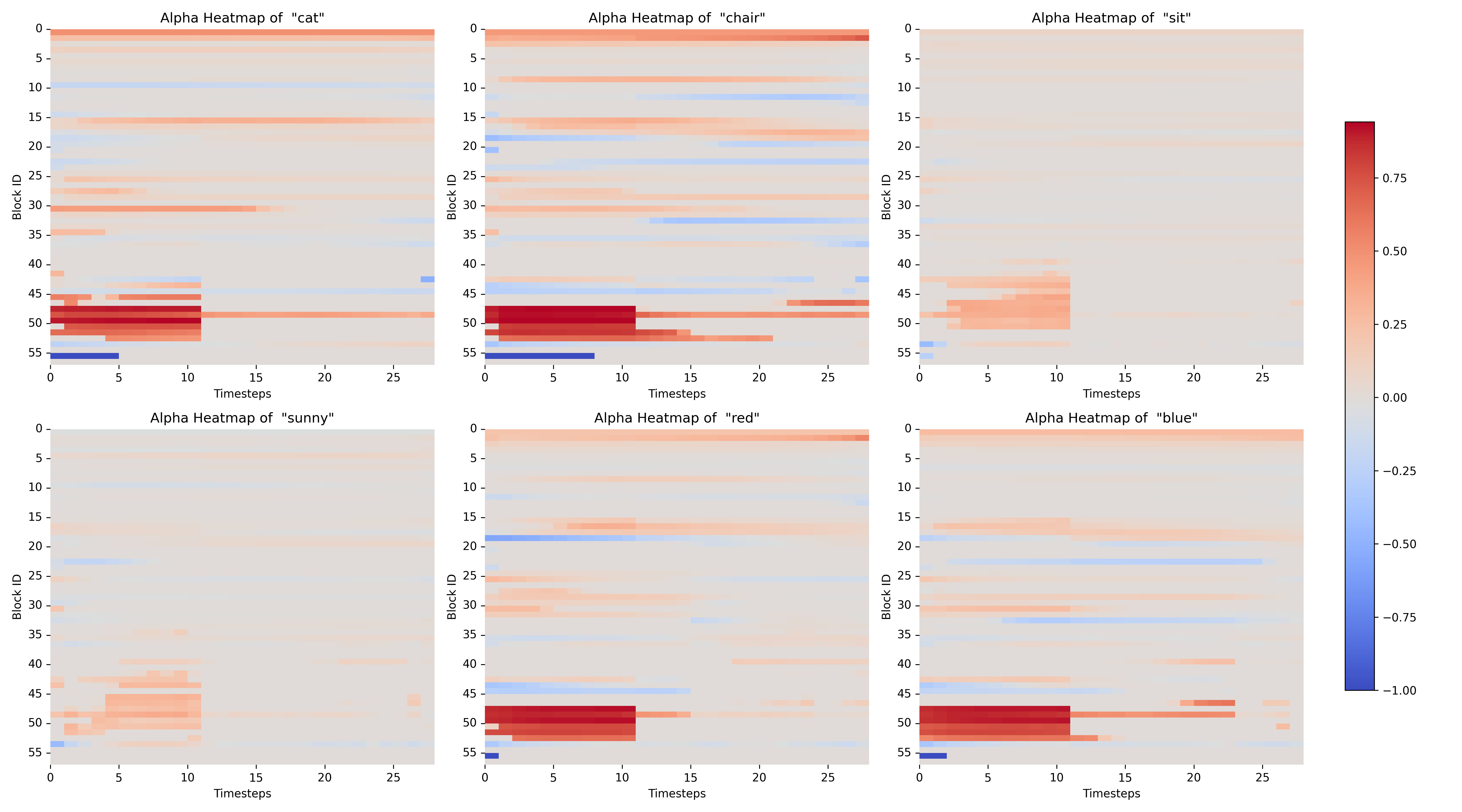}
    \vspace{-0.5em}
    \caption{
    \textbf{Per-token $\alpha$ heatmap across blocks and timesteps.}
    We visualize the adaptive weights $\alpha$ for each token in the prompt \textit{``a red cat sitting on a blue chair in a sunny garden''} across all 57 transformer blocks and 28 denoising timesteps.
    Subject tokens (\textit{``cat''}, \textit{``chair''}) show consistently high $\alpha$ values, while attribute tokens (\textit{``red''}, \textit{``blue''}) peak in early and late blocks.
    Relation tokens maintain near-zero values throughout.
    }
    \label{fig:supp_flux_heatmap_tokens}
\end{figure*}

Per-token $\alpha$ heatmaps in Fig.~\ref{fig:supp_flux_heatmap_tokens} reveal distinct modulation patterns: subject tokens (\textit{``cat''}, \textit{``chair''}) exhibit higher and more temporally stable $\alpha$ values across timesteps, while attribute tokens (\textit{``red''}, \textit{``blue''}) show $\alpha$ peaking in early blocks and later blocks (blocks 0-3, 47-53) and early timesteps.

\begin{figure*}[h]
    \centering
    \includegraphics[width=\linewidth]{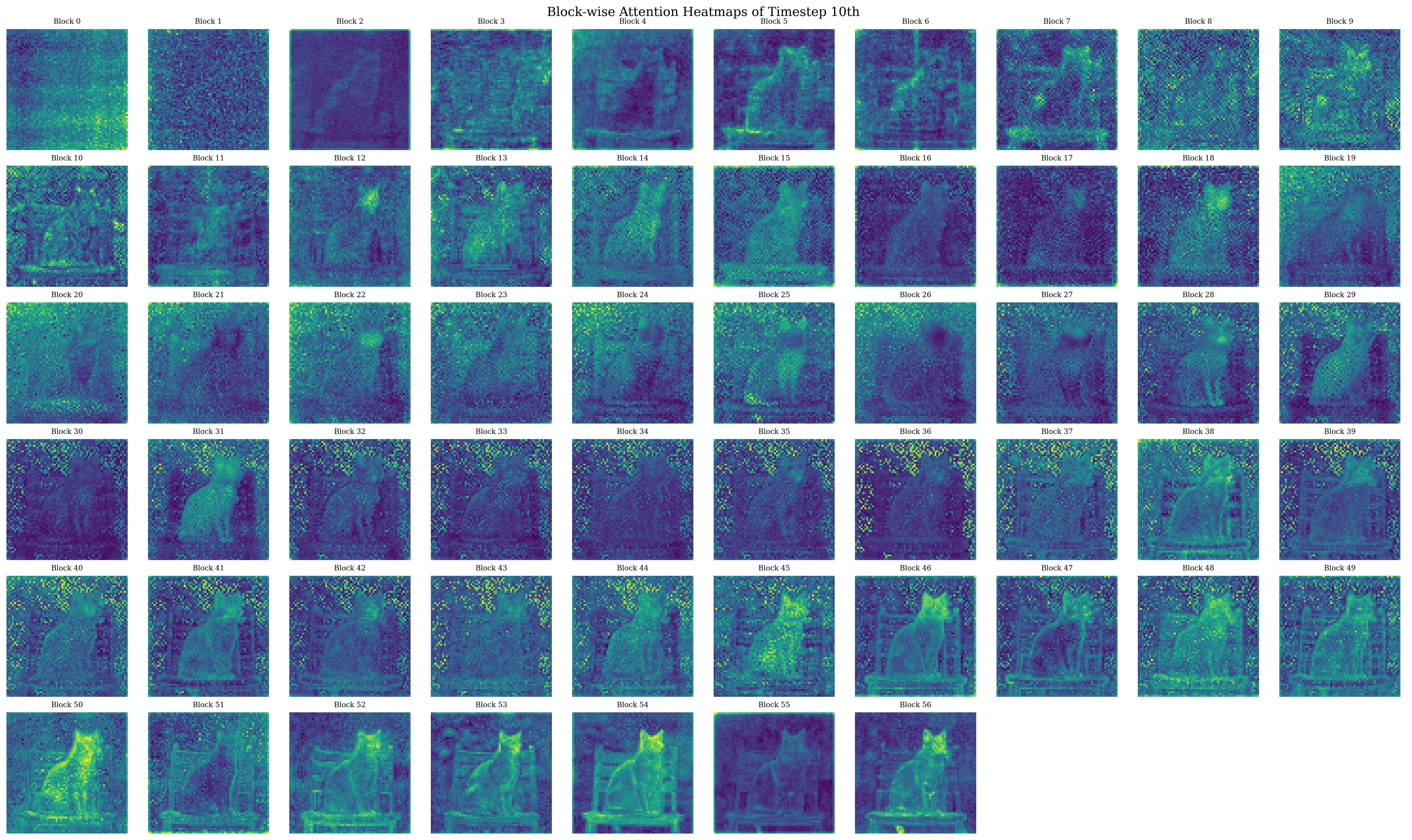}
    \vspace{-0.5em}
    \caption{
    \textbf{Block-wise spatial image-text attention maps at timestep $t=10$.}
    We visualize the spatial attention averaged across all text tokens in the prompt at different transformer blocks.
    The averaged attention is strong in early blocks (0-5) and final blocks (47-53), corresponding to regions of high $\alpha$ norm in the heatmap.
    }
    \label{fig:supp_flux_attn_blocks}
\end{figure*}

\begin{figure*}[h]
    \centering
    \includegraphics[width=\linewidth]{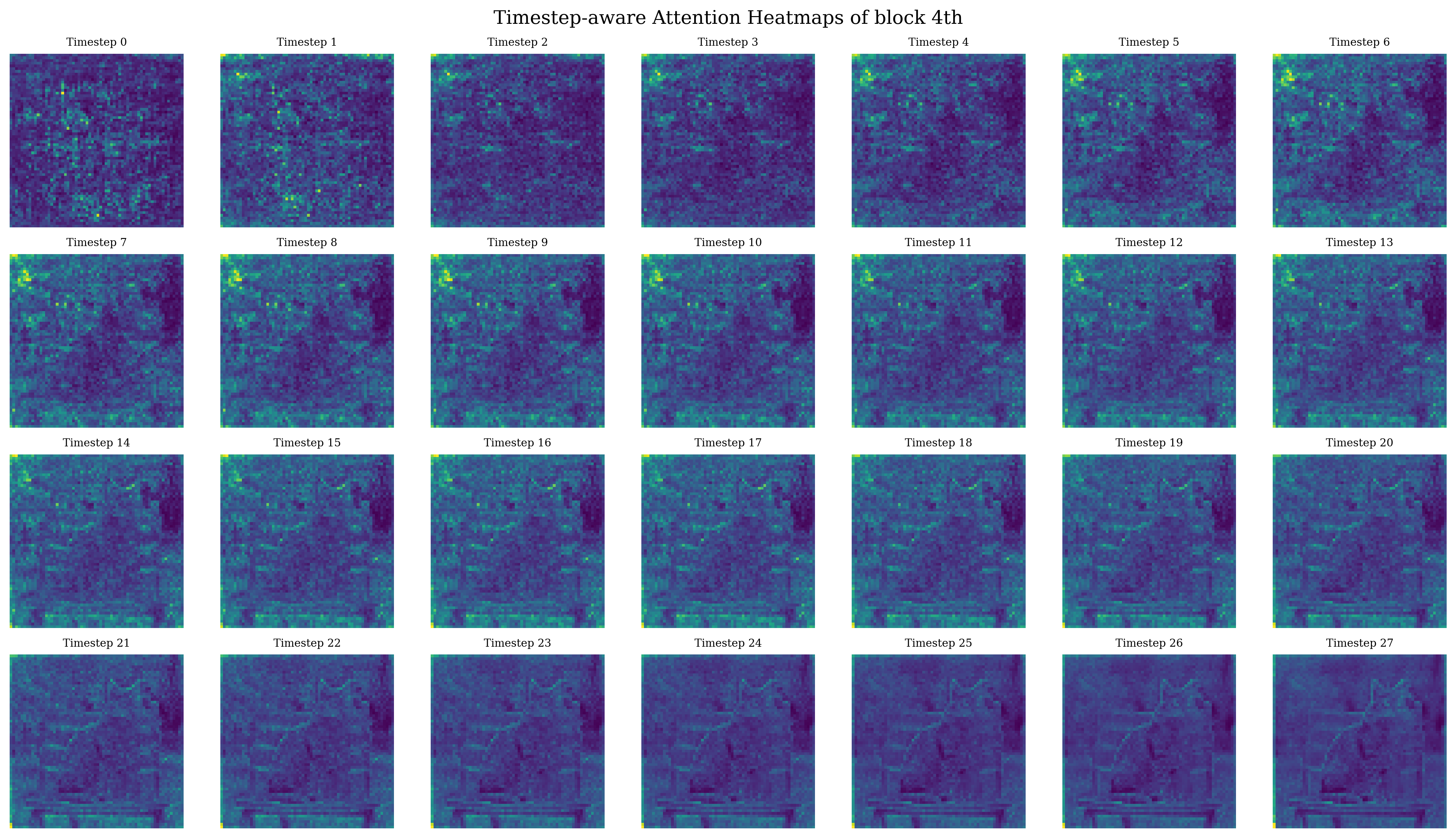}
    \vspace{-0.5em}
    \caption{
    \textbf{Timestep-wise spatial image-text attention evolution at block 4.}
    We visualize the spatial attention averaged across all text tokens in the prompt at different denoising timesteps.
    The averaged attention map evolves from diffuse (early timesteps) to spatially focused (later timesteps).
    }
    \label{fig:supp_flux_attn_timesteps}
\end{figure*}

Spatial image-text attention maps further validate this interpretability.
Block-wise visualization in Fig.~\ref{fig:supp_flux_attn_blocks} at timestep $t=10$ shows that the averaged attention across all text tokens is strong in early blocks (0-5) and final blocks (47-53), corresponding to regions of high $\alpha$ norm in the heatmap.
Early blocks establish spatial structure, while late blocks refine details.
Timestep-wise visualization in Fig.~\ref{fig:supp_flux_attn_timesteps} at block 4 shows that the averaged attention map evolves from diffuse (early timesteps) to spatially focused (later timesteps), reflecting the transition from structure generation to semantic refinement.
These fine-grained traces demonstrate that Aid's modulation is not only interpretable at the overall distribution level, but also semantically meaningful at the individual token level.

\clearpage
\newpage

\subsection{More Results on FLUX and SD 3.5}
The quantitative results below each image show the scores of HPSv3, Image Reward, and Aesthetic Score.
Our method demonstrates superior prompt adherence and image quality.

\begin{figure*}[h]
    \centering
    \includegraphics[width=0.95\linewidth]{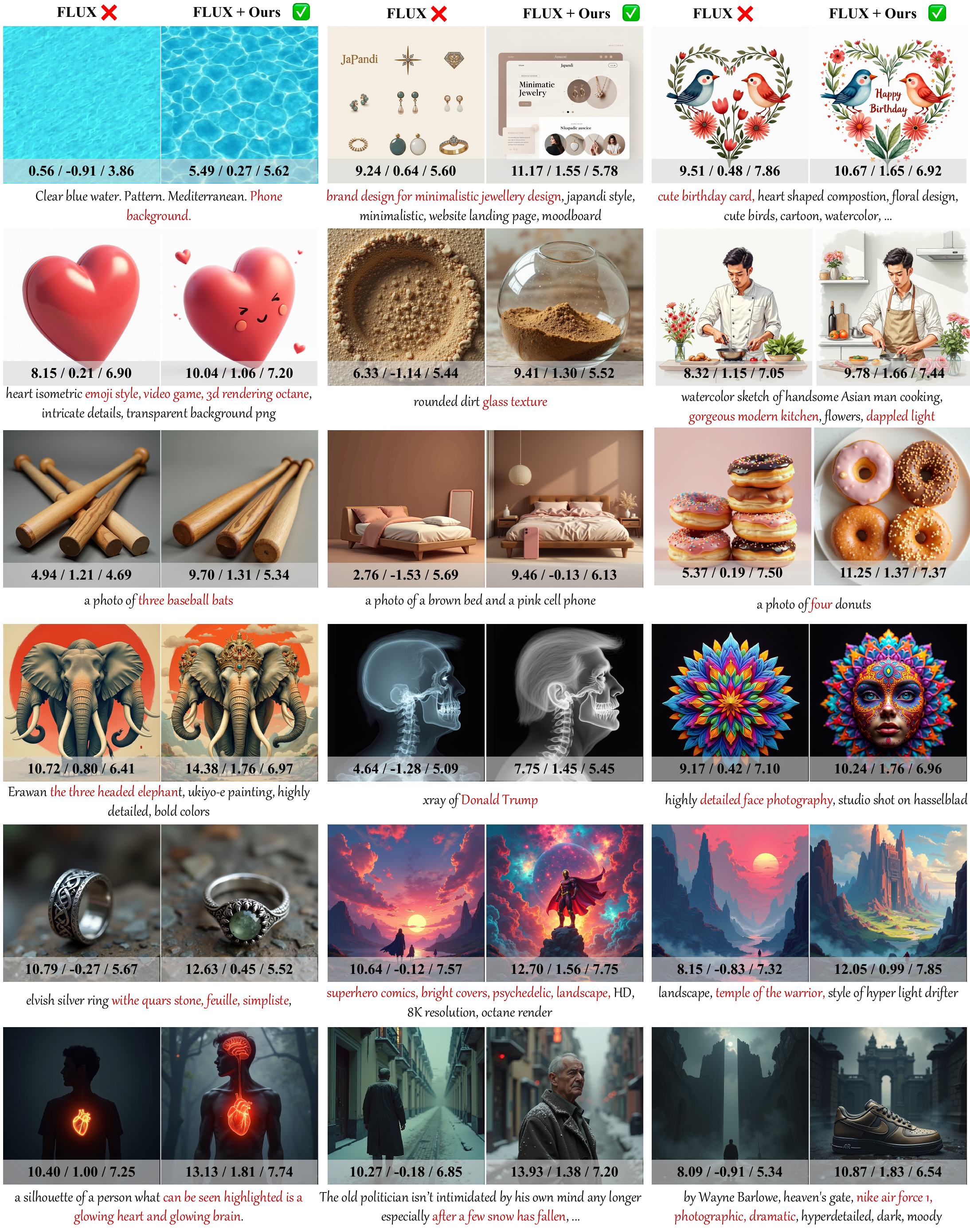}
    \vspace{-0.5em}
    \caption{
    \textbf{More Qualitative comparisons of images generated by baseline FLUX with and without our proposed \method.}
    The quantitative results below each image show the scores of HPSv3, Image Reward, and Aesthetic Score.
    Our method demonstrates superior prompt adherence and image quality.
    }
    \label{fig:supp_flux_qualitative}
\end{figure*}

\begin{figure*}[t]
    \centering
    \includegraphics[width=0.95\linewidth]{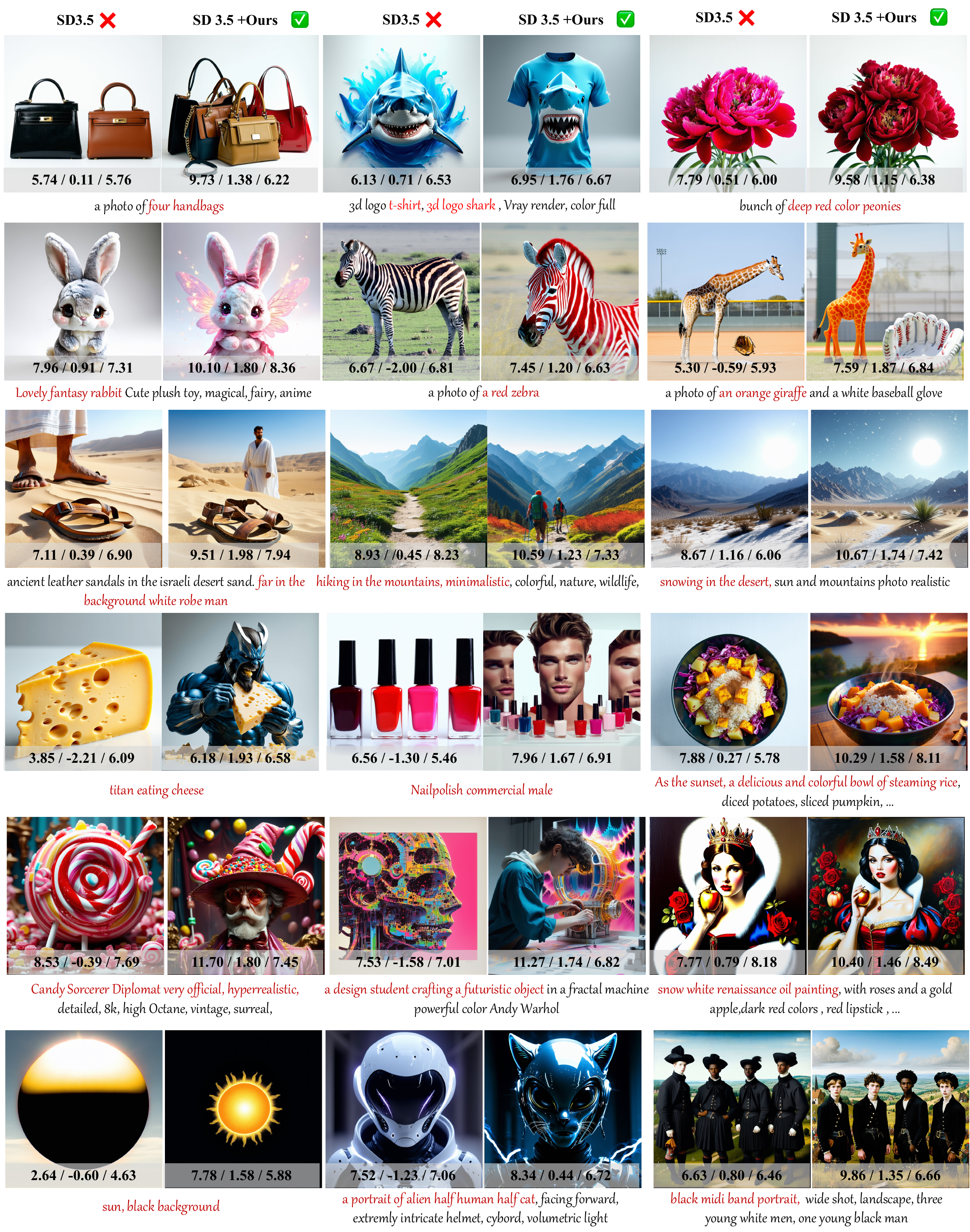}
    \vspace{-0.5em}
    \caption{
    \textbf{More Qualitative comparisons of images generated by baseline SD 3.5 with and without our proposed \method.}
    The quantitative results below each image show the scores of HPSv3, Image Reward, and Aesthetic Score.
    Our method demonstrates superior prompt adherence and image quality.
    }
    \label{fig:supp_sd3_qualitative}
\end{figure*}

\clearpage
\newpage

\subsection{Failure Cases}
\label{appendix:failure_cases}

\begin{figure*}[!htbp]
    \centering
    \vspace{1.5em}
    \includegraphics[width=0.95\linewidth]{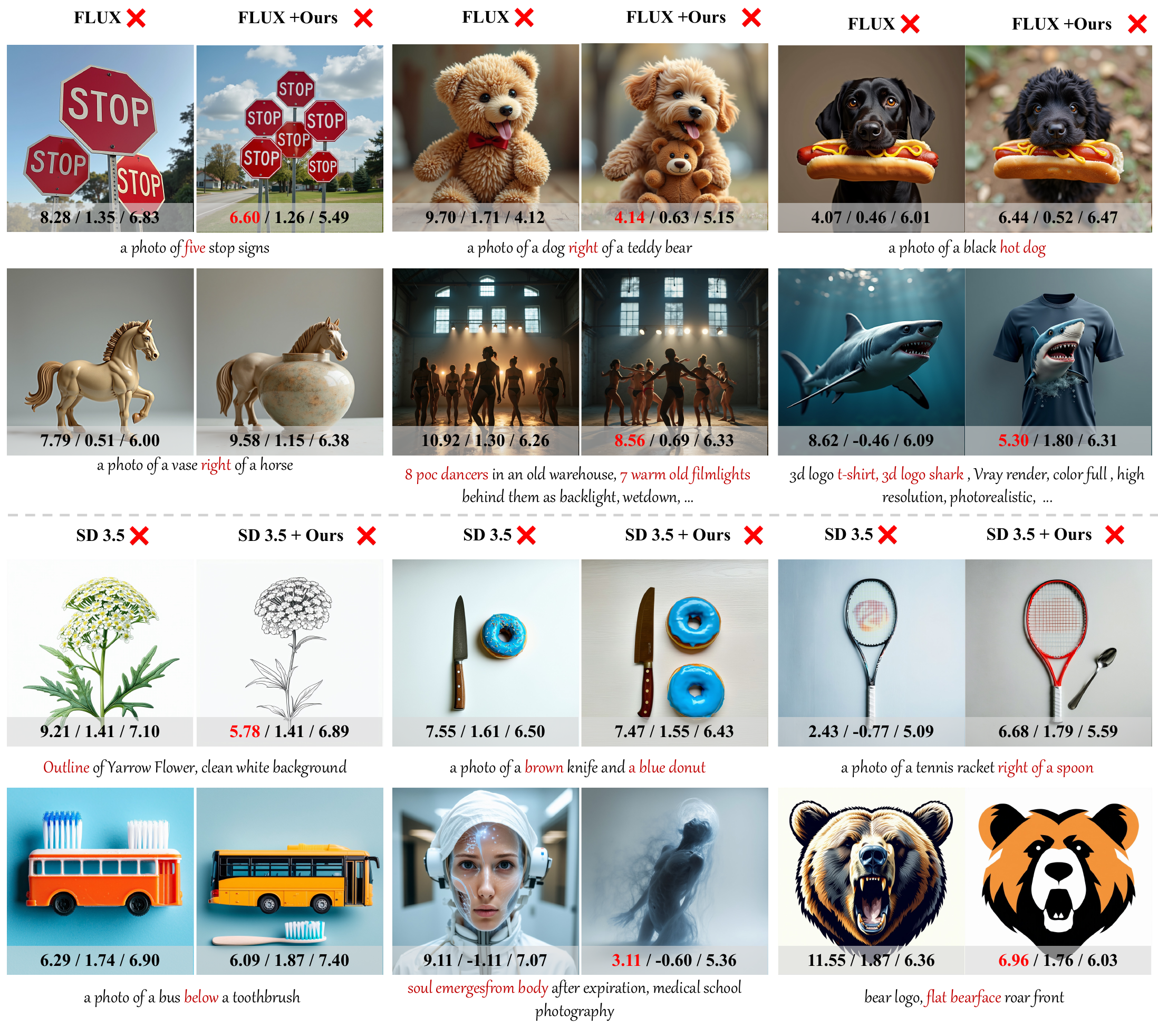}
    \caption{
    \textbf{Some failure cases of baseline models with and without our proposed \method.}
    Examples include counting and spatial relations, abstract descriptions, and rare or ambiguous concepts. Both baseline and \method can fail to satisfy the complete prompt.
    }
    \label{fig:failure_cases}
\end{figure*}

Fig.~\ref{fig:failure_cases} groups the remaining errors by what textual modulation cannot reliably supply. Counting and spatial binding remain difficult: neither method renders exactly five stop signs, and the dog, teddy bear, horse, and vase are not consistently arranged as requested. Abstract or highly constrained descriptions, such as the ``soul emerges from body'' scene and the detailed dancer prompt, leave substantial room for an incorrect visual interpretation. Rare or ambiguous concepts also expose limitations of the frozen backbone, illustrated by the yarrow flower and ``hot dog'' examples. In several cases \method recovers a requested object, but the complete relation or visual style still fails. These examples limit the claim to improved average alignment rather than guaranteed prompt satisfaction.

\newpage
\section*{NeurIPS Paper Checklist}

\begin{enumerate}

\item {\bf Claims}
    \item[] Question: Do the main claims made in the abstract and introduction accurately reflect the paper's contributions and scope?
    \item[] Answer: \answerYes{} %
    \item[] Justification: We introduce our \method in the abstract and introduction, provide a detailed description in Section \ref{sec:methods}, and present experimental results in Section \ref{sec:experiments}.
    \item[] Guidelines:
    \begin{itemize}
        \item The answer \answerNA{} means that the abstract and introduction do not include the claims made in the paper.
        \item The abstract and/or introduction should clearly state the claims made, including the contributions made in the paper and important assumptions and limitations. A \answerNo{} or \answerNA{} answer to this question will not be perceived well by the reviewers.
        \item The claims made should match theoretical and experimental results, and reflect how much the results can be expected to generalize to other settings.
        \item It is fine to include aspirational goals as motivation as long as it is clear that these goals are not attained by the paper.
    \end{itemize}

\item {\bf Limitations}
    \item[] Question: Does the paper discuss the limitations of the work performed by the authors?
    \item[] Answer: \answerYes{} %
    \item[] Justification: We discuss the limitations of this work in Section \ref{sec:conclusion}.
    \item[] Guidelines:
    \begin{itemize}
        \item The answer \answerNA{} means that the paper has no limitation while the answer \answerNo{} means that the paper has limitations, but those are not discussed in the paper.
        \item The authors are encouraged to create a separate ``Limitations'' section in their paper.
        \item The paper should point out any strong assumptions and how robust the results are to violations of these assumptions (e.g., independence assumptions, noiseless settings, model well-specification, asymptotic approximations only holding locally). The authors should reflect on how these assumptions might be violated in practice and what the implications would be.
        \item The authors should reflect on the scope of the claims made, e.g., if the approach was only tested on a few datasets or with a few runs. In general, empirical results often depend on implicit assumptions, which should be articulated.
        \item The authors should reflect on the factors that influence the performance of the approach. For example, a facial recognition algorithm may perform poorly when image resolution is low or images are taken in low lighting. Or a speech-to-text system might not be used reliably to provide closed captions for online lectures because it fails to handle technical jargon.
        \item The authors should discuss the computational efficiency of the proposed algorithms and how they scale with dataset size.
        \item If applicable, the authors should discuss possible limitations of their approach to address problems of privacy and fairness.
        \item While the authors might fear that complete honesty about limitations might be used by reviewers as grounds for rejection, a worse outcome might be that reviewers discover limitations that aren't acknowledged in the paper. The authors should use their best judgment and recognize that individual actions in favor of transparency play an important role in developing norms that preserve the integrity of the community. Reviewers will be specifically instructed to not penalize honesty concerning limitations.
    \end{itemize}

\item {\bf Theory assumptions and proofs}
    \item[] Question: For each theoretical result, does the paper provide the full set of assumptions and a complete (and correct) proof?
    \item[] Answer: \answerNA{} %
    \item[] Justification: This paper does not include theoretical results.
    \item[] Guidelines:
    \begin{itemize}
        \item The answer \answerNA{} means that the paper does not include theoretical results.
        \item All the theorems, formulas, and proofs in the paper should be numbered and cross-referenced.
        \item All assumptions should be clearly stated or referenced in the statement of any theorems.
        \item The proofs can either appear in the main paper or the supplemental material, but if they appear in the supplemental material, the authors are encouraged to provide a short proof sketch to provide intuition.
        \item Inversely, any informal proof provided in the core of the paper should be complemented by formal proofs provided in appendix or supplemental material.
        \item Theorems and Lemmas that the proof relies upon should be properly referenced.
    \end{itemize}

    \item {\bf Experimental result reproducibility}
    \item[] Question: Does the paper fully disclose all the information needed to reproduce the main experimental results of the paper to the extent that it affects the main claims and/or conclusions of the paper (regardless of whether the code and data are provided or not)?
    \item[] Answer: \answerYes{} %
    \item[] Justification: The architecture of our module is introduced in Section \ref{sec:methods}, and the implementation details are presented at the beginning of Section \ref{sec:experiments} and supplementary sections.
    \item[] Guidelines:
    \begin{itemize}
        \item The answer \answerNA{} means that the paper does not include experiments.
        \item If the paper includes experiments, a \answerNo{} answer to this question will not be perceived well by the reviewers: Making the paper reproducible is important, regardless of whether the code and data are provided or not.
        \item If the contribution is a dataset and\slash or model, the authors should describe the steps taken to make their results reproducible or verifiable.
        \item Depending on the contribution, reproducibility can be accomplished in various ways. For example, if the contribution is a novel architecture, describing the architecture fully might suffice, or if the contribution is a specific model and empirical evaluation, it may be necessary to either make it possible for others to replicate the model with the same dataset, or provide access to the model. In general. releasing code and data is often one good way to accomplish this, but reproducibility can also be provided via detailed instructions for how to replicate the results, access to a hosted model (e.g., in the case of a large language model), releasing of a model checkpoint, or other means that are appropriate to the research performed.
        \item While NeurIPS does not require releasing code, the conference does require all submissions to provide some reasonable avenue for reproducibility, which may depend on the nature of the contribution. For example
        \begin{enumerate}
            \item If the contribution is primarily a new algorithm, the paper should make it clear how to reproduce that algorithm.
            \item If the contribution is primarily a new model architecture, the paper should describe the architecture clearly and fully.
            \item If the contribution is a new model (e.g., a large language model), then there should either be a way to access this model for reproducing the results or a way to reproduce the model (e.g., with an open-source dataset or instructions for how to construct the dataset).
            \item We recognize that reproducibility may be tricky in some cases, in which case authors are welcome to describe the particular way they provide for reproducibility. In the case of closed-source models, it may be that access to the model is limited in some way (e.g., to registered users), but it should be possible for other researchers to have some path to reproducing or verifying the results.
        \end{enumerate}
    \end{itemize}

\item {\bf Open access to data and code}
    \item[] Question: Does the paper provide open access to the data and code, with sufficient instructions to faithfully reproduce the main experimental results, as described in supplemental material?
    \item[] Answer: \answerNo{} %
    \item[] Justification: We will release our code upon acceptance.
    \item[] Guidelines:
    \begin{itemize}
        \item The answer \answerNA{} means that paper does not include experiments requiring code.
        \item Please see the NeurIPS code and data submission guidelines (\url{https://neurips.cc/public/guides/CodeSubmissionPolicy}) for more details.
        \item While we encourage the release of code and data, we understand that this might not be possible, so \answerNo{} is an acceptable answer. Papers cannot be rejected simply for not including code, unless this is central to the contribution (e.g., for a new open-source benchmark).
        \item The instructions should contain the exact command and environment needed to run to reproduce the results. See the NeurIPS code and data submission guidelines (\url{https://neurips.cc/public/guides/CodeSubmissionPolicy}) for more details.
        \item The authors should provide instructions on data access and preparation, including how to access the raw data, preprocessed data, intermediate data, and generated data, etc.
        \item The authors should provide scripts to reproduce all experimental results for the new proposed method and baselines. If only a subset of experiments are reproducible, they should state which ones are omitted from the script and why.
        \item At submission time, to preserve anonymity, the authors should release anonymized versions (if applicable).
        \item Providing as much information as possible in supplemental material (appended to the paper) is recommended, but including URLs to data and code is permitted.
    \end{itemize}

\item {\bf Experimental setting/details}
    \item[] Question: Does the paper specify all the training and test details (e.g., data splits, hyperparameters, how they were chosen, type of optimizer) necessary to understand the results?
    \item[] Answer: \answerYes{} %
    \item[] Justification: The implementation details are presented at the beginning of Section \ref{sec:experiments}.
    \item[] Guidelines:
    \begin{itemize}
        \item The answer \answerNA{} means that the paper does not include experiments.
        \item The experimental setting should be presented in the core of the paper to a level of detail that is necessary to appreciate the results and make sense of them.
        \item The full details can be provided either with the code, in appendix, or as supplemental material.
    \end{itemize}

\item {\bf Experiment statistical significance}
    \item[] Question: Does the paper report error bars suitably and correctly defined or other appropriate information about the statistical significance of the experiments?
    \item[] Answer: \answerNo{} %
    \item[] Justification: Although we report multiple evaluation metrics (e.g., HPSv2, HPSv3, ImageReward) showing significant improvements over baselines, we did not include error bars, standard deviations, confidence intervals, or statistical significance tests across multiple runs because it would be too computationally expensive. As such, our results do not fully meet the criteria outlined in this question for statistical rigor and uncertainty quantification.
    \item[] Guidelines:
    \begin{itemize}
        \item The answer \answerNA{} means that the paper does not include experiments.
        \item The authors should answer \answerYes{} if the results are accompanied by error bars, confidence intervals, or statistical significance tests, at least for the experiments that support the main claims of the paper.
        \item The factors of variability that the error bars are capturing should be clearly stated (for example, train/test split, initialization, random drawing of some parameter, or overall run with given experimental conditions).
        \item The method for calculating the error bars should be explained (closed form formula, call to a library function, bootstrap, etc.)
        \item The assumptions made should be given (e.g., Normally distributed errors).
        \item It should be clear whether the error bar is the standard deviation or the standard error of the mean.
        \item It is OK to report 1-sigma error bars, but one should state it. The authors should preferably report a 2-sigma error bar than state that they have a 96\% CI, if the hypothesis of Normality of errors is not verified.
        \item For asymmetric distributions, the authors should be careful not to show in tables or figures symmetric error bars that would yield results that are out of range (e.g., negative error rates).
        \item If error bars are reported in tables or plots, the authors should explain in the text how they were calculated and reference the corresponding figures or tables in the text.
    \end{itemize}

\item {\bf Experiments compute resources}
    \item[] Question: For each experiment, does the paper provide sufficient information on the computer resources (type of compute workers, memory, time of execution) needed to reproduce the experiments?
    \item[] Answer: \answerYes{} %
    \item[] Justification: We present the computer resources needed for the experiments at the beginning of Section \ref{sec:experiments}, and time of execution in supplementary sections.
    \item[] Guidelines:
    \begin{itemize}
        \item The answer \answerNA{} means that the paper does not include experiments.
        \item The paper should indicate the type of compute workers CPU or GPU, internal cluster, or cloud provider, including relevant memory and storage.
        \item The paper should provide the amount of compute required for each of the individual experimental runs as well as estimate the total compute.
        \item The paper should disclose whether the full research project required more compute than the experiments reported in the paper (e.g., preliminary or failed experiments that didn't make it into the paper).
    \end{itemize}

\item {\bf Code of ethics}
    \item[] Question: Does the research conducted in the paper conform, in every respect, with the NeurIPS Code of Ethics \url{https://neurips.cc/public/EthicsGuidelines}?
    \item[] Answer: \answerYes{} %
    \item[] Justification: The research were conducted in the paper conform, in every respect, with the NeurIPS Code of Ethics.
    \item[] Guidelines:
    \begin{itemize}
        \item The answer \answerNA{} means that the authors have not reviewed the NeurIPS Code of Ethics.
        \item If the authors answer \answerNo, they should explain the special circumstances that require a deviation from the Code of Ethics.
        \item The authors should make sure to preserve anonymity (e.g., if there is a special consideration due to laws or regulations in their jurisdiction).
    \end{itemize}

\item {\bf Broader impacts}
    \item[] Question: Does the paper discuss both potential positive societal impacts and negative societal impacts of the work performed?
    \item[] Answer: \answerYes{} %
    \item[] Justification: We discuss the broader impact in section \ref{sec:conclusion}.
    \item[] Guidelines:
    \begin{itemize}
        \item The answer \answerNA{} means that there is no societal impact of the work performed.
        \item If the authors answer \answerNA{} or \answerNo, they should explain why their work has no societal impact or why the paper does not address societal impact.
        \item Examples of negative societal impacts include potential malicious or unintended uses (e.g., disinformation, generating fake profiles, surveillance), fairness considerations (e.g., deployment of technologies that could make decisions that unfairly impact specific groups), privacy considerations, and security considerations.
        \item The conference expects that many papers will be foundational research and not tied to particular applications, let alone deployments. However, if there is a direct path to any negative applications, the authors should point it out. For example, it is legitimate to point out that an improvement in the quality of generative models could be used to generate Deepfakes for disinformation. On the other hand, it is not needed to point out that a generic algorithm for optimizing neural networks could enable people to train models that generate Deepfakes faster.
        \item The authors should consider possible harms that could arise when the technology is being used as intended and functioning correctly, harms that could arise when the technology is being used as intended but gives incorrect results, and harms following from (intentional or unintentional) misuse of the technology.
        \item If there are negative societal impacts, the authors could also discuss possible mitigation strategies (e.g., gated release of models, providing defenses in addition to attacks, mechanisms for monitoring misuse, mechanisms to monitor how a system learns from feedback over time, improving the efficiency and accessibility of ML).
    \end{itemize}

\item {\bf Safeguards}
    \item[] Question: Does the paper describe safeguards that have been put in place for responsible release of data or models that have a high risk for misuse (e.g., pre-trained language models, image generators, or scraped datasets)?
    \item[] Answer: \answerYes{} %
    \item[] Justification: We use the existing excellent SD3.5, Flux.1 and their safeguards. Our model will not bring more risk than SD3.5 or Flux.1.
    \item[] Guidelines:
    \begin{itemize}
        \item The answer \answerNA{} means that the paper poses no such risks.
        \item Released models that have a high risk for misuse or dual-use should be released with necessary safeguards to allow for controlled use of the model, for example by requiring that users adhere to usage guidelines or restrictions to access the model or implementing safety filters.
        \item Datasets that have been scraped from the Internet could pose safety risks. The authors should describe how they avoided releasing unsafe images.
        \item We recognize that providing effective safeguards is challenging, and many papers do not require this, but we encourage authors to take this into account and make a best faith effort.
    \end{itemize}

\item {\bf Licenses for existing assets}
    \item[] Question: Are the creators or original owners of assets (e.g., code, data, models), used in the paper, properly credited and are the license and terms of use explicitly mentioned and properly respected?
    \item[] Answer: \answerYes{} %
    \item[] Justification: We work with open-source models that are publicly available, and we cited them properly. Especially, FLUX.1-Dev uses the \href{https://github.com/black-forest-labs/flux/blob/main/model_licenses/LICENSE-FLUX1-dev}{Non-Commercial License}.
    \item[] Guidelines:
    \begin{itemize}
        \item The answer \answerNA{} means that the paper does not use existing assets.
        \item The authors should cite the original paper that produced the code package or dataset.
        \item The authors should state which version of the asset is used and, if possible, include a URL.
        \item The name of the license (e.g., CC-BY 4.0) should be included for each asset.
        \item For scraped data from a particular source (e.g., website), the copyright and terms of service of that source should be provided.
        \item If assets are released, the license, copyright information, and terms of use in the package should be provided. For popular datasets, \url{paperswithcode.com/datasets} has curated licenses for some datasets. Their licensing guide can help determine the license of a dataset.
        \item For existing datasets that are re-packaged, both the original license and the license of the derived asset (if it has changed) should be provided.
        \item If this information is not available online, the authors are encouraged to reach out to the asset's creators.
    \end{itemize}

\item {\bf New assets}
    \item[] Question: Are new assets introduced in the paper well documented and is the documentation provided alongside the assets?
    \item[] Answer: \answerYes{} %
    \item[] Justification: The dataset defined for the model training will be made publicly available, in case of acceptance, together with the documentation required for reproducing the experiments. Moreover, in case of acceptance, we will also release the source code of our model.
    \item[] Guidelines:
    \begin{itemize}
        \item The answer \answerNA{} means that the paper does not release new assets.
        \item Researchers should communicate the details of the dataset\slash code\slash model as part of their submissions via structured templates. This includes details about training, license, limitations, etc.
        \item The paper should discuss whether and how consent was obtained from people whose asset is used.
        \item At submission time, remember to anonymize your assets (if applicable). You can either create an anonymized URL or include an anonymized zip file.
    \end{itemize}

\item {\bf Crowdsourcing and research with human subjects}
    \item[] Question: For crowdsourcing experiments and research with human subjects, does the paper include the full text of instructions given to participants and screenshots, if applicable, as well as details about compensation (if any)?
    \item[] Answer: \answerYes{} %
    \item[] Justification: We recruited 8 volunteers for the blinded pairwise evaluation; the voting protocol is described in the main text.
    \item[] Guidelines:
    \begin{itemize}
        \item The answer \answerNA{} means that the paper does not involve crowdsourcing nor research with human subjects.
        \item Including this information in the supplemental material is fine, but if the main contribution of the paper involves human subjects, then as much detail as possible should be included in the main paper.
        \item According to the NeurIPS Code of Ethics, workers involved in data collection, curation, or other labor should be paid at least the minimum wage in the country of the data collector.
    \end{itemize}

\item {\bf Institutional review board (IRB) approvals or equivalent for research with human subjects}
    \item[] Question: Does the paper describe potential risks incurred by study participants, whether such risks were disclosed to the subjects, and whether Institutional Review Board (IRB) approvals (or an equivalent approval/review based on the requirements of your country or institution) were obtained?
    \item[] Answer: \answerNA{} %
    \item[] Justification: This paper does not involve crowdsourcing nor research with human subjects.
    \item[] Guidelines:
    \begin{itemize}
        \item The answer \answerNA{} means that the paper does not involve crowdsourcing nor research with human subjects.
        \item Depending on the country in which research is conducted, IRB approval (or equivalent) may be required for any human subjects research. If you obtained IRB approval, you should clearly state this in the paper.
        \item We recognize that the procedures for this may vary significantly between institutions and locations, and we expect authors to adhere to the NeurIPS Code of Ethics and the guidelines for their institution.
        \item For initial submissions, do not include any information that would break anonymity (if applicable), such as the institution conducting the review.
    \end{itemize}

\item {\bf Declaration of LLM usage}
    \item[] Question: Does the paper describe the usage of LLMs if it is an important, original, or non-standard component of the core methods in this research? Note that if the LLM is used only for writing, editing, or formatting purposes and does \emph{not} impact the core methodology, scientific rigor, or originality of the research, declaration is not required.
    \item[] Answer: \answerNA{} %
    \item[] Justification: The core method development in this research does not involve LLMs as any important, original, or non-standard components.
    \item[] Guidelines:
    \begin{itemize}
        \item The answer \answerNA{} means that the core method development in this research does not involve LLMs as any important, original, or non-standard components.
        \item Please refer to our LLM policy in the NeurIPS handbook for what should or should not be described.
    \end{itemize}

\end{enumerate}

\end{document}